\documentclass{article}

\usepackage[dvipsnames]{xcolor}

\usepackage{microtype}
\usepackage{graphicx}
\usepackage{booktabs} %
\usepackage{subcaption}
\usepackage{multirow}
\usepackage{caption}
\usepackage{amsmath,amsfonts,bm}

\def\eqref#1{equation~\ref{#1}}

\def\1{\bm{1}}

\newcommand\norm[2]{\left|\!\left|#1 \right|\!\right|_{#2}}
\newcommand{\nn}{\mathcal{F}}
\newcommand{\x}{\mathbf{x}}
\newcommand{\y}{\mathbf{y}}

\newcommand{\dataset}{\mathcal{S}}

\DeclareMathAlphabet{\mathsfit}{\encodingdefault}{\sfdefault}{m}{sl}
\SetMathAlphabet{\mathsfit}{bold}{\encodingdefault}{\sfdefault}{bx}{n}

\usepackage{floatrow}

\usepackage{hyperref}

\usepackage[accepted]{icml2023}

\usepackage{amsmath}
\usepackage{amssymb}
\usepackage{mathtools}
\usepackage{amsthm}
\usepackage{nicefrac}

\usepackage[capitalize,noabbrev]{cleveref}

\theoremstyle{plain}

\theoremstyle{definition}

\theoremstyle{remark}

\usepackage[textsize=tiny]{todonotes}

\icmltitlerunning{Can Neural Network Memorization Be Localized?}
\begin{document}

\twocolumn[
\icmltitle{Can Neural Network Memorization Be Localized?}

\icmlsetsymbol{equal}{*}

\begin{icmlauthorlist}
\icmlauthor{Pratyush Maini}{cmu}
\icmlauthor{Michael C. Mozer}{goo}
\icmlauthor{Hanie Sedghi}{goo}
\icmlauthor{Zachary C. Lipton}{cmu}
\icmlauthor{J. Zico Kolter}{cmu}
\icmlauthor{Chiyuan Zhang}{goo}
\end{icmlauthorlist}

\icmlaffiliation{cmu}{School of Computer Science, Carnegie Mellon University}
\icmlaffiliation{goo}{Google Research, Mountain View, CA}

\icmlcorrespondingauthor{Pratyush Maini}{pratyushmaini@cmu.edu}

\icmlkeywords{Machine Learning, ICML}

\vskip 0.3in
]

\printAffiliationsAndNotice{}  %

\begin{abstract}
  Recent efforts at explaining the interplay of memorization and generalization in deep overparametrized networks have posited that neural networks \emph{memorize} ``hard'' examples in the final few layers of the model. Memorization refers to the ability to correctly predict on \emph{atypical} examples of the training set. 
  In this work, we 
  show that rather than being confined to individual layers,  memorization is a phenomenon confined to a small set of neurons in various layers of the model.
  First, via three experimental sources of converging evidence, we find that most layers are redundant for the memorization 
  of examples and the layers that contribute to example memorization are, in general, not the final layers.
  The three sources are \emph{gradient accounting} (measuring the contribution to the gradient norms from memorized and 
  clean examples), \emph{layer rewinding} (replacing specific model weights of a converged model with previous 
  training checkpoints), and \emph{retraining} (training rewound layers only on clean examples).
  Second, we ask a more generic question: can memorization be localized \emph{anywhere} in a model?
  We discover that 
  memorization is often confined to a small number of neurons or channels (around 5) of the model.  
  Based on these insights we propose a new form of dropout---\emph{example-tied dropout} that enables us to direct the memorization of examples to an a priori determined set of neurons. 
  By dropping out these  neurons, 
  we are able to reduce the accuracy on memorized examples from $100\%\to3\%$, while also reducing the generalization gap.\footnote{Code for reproducing our experiments can be found at \href{https://github.com/pratyushmaini/localizing-memorization}{https://github.com/pratyushmaini/localizing-memorization}.}

  \end{abstract}

\section{Introduction}
Deep neural networks 
are capable of memorizing randomly labeled data~\citep{zhang2017understanding} yet
generalize remarkably well on 
real world
data. Theories have been developed to %
understand this puzzling phenomenon.
For example, \emph{implicit regularization} argues that the training algorithms, i.e., SGD and its variants, implicitly encourage the trained model to converge to generalizable solutions when the underlying data distribution is well-behaved \citep{soudry2018implicit,smith2021on}. Alternatively, the theory of \emph{benign overfitting} argues that when a neural network is sufficiently overparameterized, 
it generalizes to the majority of the dataset using a \emph{simple} component of the input, and memorizes mislabeled and irregular data using a \emph{complex} component of the input data~\citep{bartlett2020benign}. These components do not interfere, making such overfitting benign.
Another theory of label memorization suggests that when the data follows a long-tailed distribution, memorizing rare examples in the tail is necessary for achieving optimal generalization~\citep{feldman2020does,feldman2020neural}. 
One common notion of memorization is the `leave-one-out' memorization---does the prediction of an (atypical) example change significantly if it was removed from the training set? Since this is computationally expensive, multiple prior works operationalize memorization via mislabeled examples, which \emph{must} be memorized for correct prediction. In our study, we consider both types---mislabeled and atypical example memorization.

The practical implication of memorization in deep neural networks has also been widely studied, especially from a privacy and security perspective. For example, \citet{carlini2021extraction} shows that it is possible to extract training data from large language models. 
Other work has attempted to reconstruct training examples from output activations~\citep{teterwak2021understanding} or even parameters~\citep{haim2022reconstructing} of \emph{discriminatively} trained neural networks.
However, previous studies have largely focused on the implications of memorization and/or identifying memorized examples. The question of \emph{where memorization happens} in a  neural network remains unclear with only anecdotal observations. 
In particular, it is \emph{commonly} believed that memorization of training examples happens in the last (or final few) layers of a deep network \citep{baldock2021deep,stephenson21geometry}. 

In this work, we show that memorization is confined to a small subset of neurons distributed across all layers, as opposed to a fixed set of (such as the final few) layers.
We conduct a systematic study to understand the mechanism of memorization in neural networks, in particular, whether there are specific locations in which this phenomenon materializes. Then, we propose a simple dropout modification that enables directing memorization to a fixed set of network neurons.
We perform our investigation in two stages. 

In Stage I, 
to verify previous beliefs that memorization is confined to the last (few) layers,
we design two experiments to detect memorization at the layer level. 
First,
by measuring the influence of memorized (noisy) examples on model weights via gradient accounting (Section~\ref{sec:gradient-accounting}) we make two key discoveries: (i) the average contribution to gradient norms by memorized examples is an order of magnitude larger than that by typical (clean) examples; and (ii) gradient contribution from the two groups of examples (clean and noisy) have an extremely negative cosine similarity throughout training, at every layer of the model. This suggests that at a layer level, the learning of noisy examples opposes the learning of clean examples, and indicates that mislabeled example overfitting may not be benign in practical settings. Interestingly, the last layer, or any other individual layer, does not stand out as receiving more influence from the memorized examples than from the typical examples.
Second, to understand the functional influence of these weight updates, we draw inferences via weight rewinding and retraining (Section~\ref{sec:functional}). In \emph{weight rewinding}, 
we replace weights of individual layers of the converged model with earlier training checkpoints, and measure the accuracy on the training set. We find that the final layers are not critical to the correct prediction of noisy examples. Moreover, as the model and problem complexity  change, we find that the critical layers of the model begin to change, suggesting the critical layers, if any, may be dependent on the particular problem instance. 
To build on the experiment of layer rewinding, keeping the rest of the model frozen, we retrain the re-initialized layers only on clean examples, and make a surprising discovery---for most layers (including the last few layers), training on clean examples confers nearly 100\% accuracy on the (now unseen) mislabeled examples---i.e., the model predicts the randomly assigned (incorrect) label even without seeing that example, indicating the redundance of these layers for memorization.

In Stage II, we move away from the hypothesis of layer-level example memorization and inspect memorization at a neuron level, akin to the ``grandmother cell'' of \citet{Barlow1972} (Section~\ref{sec:neuron-level-memorization}). By iteratively removing important neurons with respect to a sample, we determine the minimum number of neurons required to flip an example's predictions. We find that memorized (noisy) examples on average require significantly fewer neurons (by a factor of $1/3$) for flipping the predictions, as opposed to clean examples. Moreover, we make the observation that the neurons chosen as most important by our greedy search algorithm are scattered over multiple layers of the model, and the distribution of such neurons for both clean and mislabeled examples are similar.

Inspired by this observation, we
develop a training mechanism that enables us to intentionally direct the memorization of examples 
to a predetermined set of neurons. 
Specifically, we propose a new form of dropout that we call \emph{example-tied dropout} (Section~\ref{subsec:nested-dropout}).  
We keep a fraction of (generalization) neurons always active during training. A complimentary small fraction of example-specific (memorization) neurons are only activated  when that example is sampled in a mini-batch. Using this training scheme, we are able to strongly concentrate the memorization of examples to the memorization neurons. In particular, dropping out these neurons during evaluation brings down the accuracy on mislabeled examples from 100\% to 0.1\% in the case of the MNIST dataset and from 99\% to 3\% on the CIFAR10 dataset, with a minor impact on the accuracy on clean training examples. Moreover, we note that the clean examples that get misclassified upon dropping out of the memorization neurons are in fact atypical or ambiguous examples.

In summary, our findings suggest that memorization of examples by neural networks can not be localized to the last, or any few layers, but is often determined by a small set of neurons that may be distributed across layers. Moreover, we can perform training time modifications to restrict the model's memorization to a predefined set of neurons.

\section{Related Work}
\textbf{Learning mechanism.}
The learning dynamics of neural networks and their interplay with properties or characteristics of the underlying data is central to various schools of thought---(i) simplicity bias~\citep{shah2020pitfalls, arpit2017closer}, neural networks have a strong bias towards learning `simpler' features as compared to ones that require a more complex representation; (ii) long-tailed nature of datasets~\citep{feldman2020does, feldman2020neural}---
 ML datasets have a large fraction of singleton examples with features that occur only once in the training set. As a result, the neural network must memorize these examples to be able to correctly predict them; and (iii) early learning phenomenon~\citep{Frankle2020The,liu2020early}, suggesting that simpler examples are learned fast.  

\textbf{Atypical Example Memorization.}
Inspired by the early learning phenomenon,
recent work has used metrics inspired by the learning time of an example to determine which examples are memorized by a neural network~\cite{carlini2019distribution,jiang2021}. Other works have focussed on notions of forgetting time, suggesting that examples forgotten frequently during training~\cite{toneva2018an} or forgotten quickly when held out~\cite{maini2022characterizing} are memorized.

\textbf{Location of Memorization.}
Motivating our inquiry, recent works have argued that hard examples are learned in the last few layers of a model. \citet{baldock2021deep} investigate the \emph{prediction depth} of various examples, defined as the earliest layer in a neural network, whose representations can successfully predict the label of an example when probed using a k-nearest neighbor search. They find that the prediction of mislabeled examples occurs in the final few layers of the model, and conclude that ``early layers generalize while later layers
memorize''.
Other works have attempted at measuring the manifold complexity and shattering capability~\citep{stephenson21geometry} of various layers to once again argue that mislabeled examples must be memorized in the last few layers of the model. They also propose regularization of the final layer of the model as a remedy for alleviating memorization.

\textbf{Modifying neural networks.}
Recent works have attempted to change the neural network predictions by modifying only a small fraction of the neurons. Such investigations have been primarily focused on understanding where are facts stored in a neural network~\cite{zhu2021modifying,meng2022locating}. \citet{Sinitsin2020Editable} investigated methods for patching mistakes of a neural network by modifying a small number of neurons. In our experiments on neuron-level memorization, we investigate the opposite idea, what are the minimum number of neurons required to flip the prediction of a neural network to a \emph{different} class, and contrast the model behaviour on clean and memorized examples based on this. 

\textbf{Task specific neurons.}
Prior work in the space of multi-task learning has studied the use of designated gates for the task of continual learning to avoid forgetting~\cite{jacobs1991adaptive,aljundi2017expert}. Follow-up work also studied the idea of specialization at a synapse level, rather than having individual gates for different tasks~\cite{zenke2017continual,cheung2019superposition}. Recent work on training under noisy data assigned a new parameter to each example that learns the correct label for that example. This learning adapts the class label simultaneously as the model learns to predict correctly on the adapted label~\cite{pmlr-v162-liu22w}. These ideas are related to the concept of example-tied dropout that we introduce in Section~\ref{subsec:nested-dropout}. In our work, we propose a new dropout scheme that aims to select specialized memorization neurons and use it to direct all example memorization to them. This idea of specialization is akin to that in multitask learning but in a totally different, example-level regime.

\section{Problem Setup}
\label{sec:prelim}
\textbf{Preliminaries.}
Consider a $d$-layer feed-forward neural network
$\nn_d$ parametrized by weights $\theta = \left(\theta_1,\dots,\theta_d\right)$.   We use $z_l$ to denote the activations from neurons in the $l^\text{th}$ layer of the model, and refer to the equations of the network as:
\begin{equation}
z_{l} = \nn^l(z_{l-1}; \theta_l)
\end{equation}
where $\nn^l$ denotes the function of the $l^\text{th}$ layer.  When referring to a single neuron (or in the case of a convolutional network, to a single channel) we use the notation $[z_l]_j$ to refer to the $j^\text{th}$ individual neuron/channel.
We consider the supervised classification paradigm where we train 
$\nn_d$ 
on a dataset 
$\dataset = \{\x_i,\y_i\}^n$. 
The model is trained using Stochastic Gradient Descent (SGD) to minimize the empirical risk given by $\mathcal{L}(\dataset,\nn_d) = \sum_i\ell(\nn_d(\x_i),\y_i)$. We denote the model parameters after the $t^\text{th}$ epoch of training by $\theta^t = \left(\theta_1^t,\dots,\theta_d^t\right)$.

For experiments with label noise, we randomly change the label of a fixed fraction of examples to an incorrect label. $\dataset = \dataset_c \bigcup \dataset_n$, where $\dataset_c$ and $\dataset_n$ are mutually exclusive subsets containing clean and mislabeled examples respectively.

\textbf{Datasets and Models Used.}
We perform experiments on three image classification datasets, CIFAR-10~\citep{Krizhevsky09learningmultiple}, MNIST~\cite{deng2012mnist}, and SVHN~\cite{netzer2011svhn}. We use two sizes of the ResNet model, namely, ResNet-9, and ResNet-50~\citep{he2016deep}, and also the Vision Transformer (ViT) small model~\citep{dosovitskiy2020image}. Results for gradient accounting and weight rewinding of ResNet models are averaged over 3 seeds.

\textbf{Combining Layers.}
For our analysis of ResNet-50 and ViT models, we combine the influence of multiple layers of the model to simplify our visualizations. This is done by combining all layers that belong to the same block in the case of ResNet-50. The blocks are differentiated by the image size that they encode and can be identified by the layer name (\texttt{layerX.[module]}) in the standard PyTorch implementation. Similarly, for ViT, we group multiple layers into a single module based on the transformer blocks, which can be identified by the layer name (\texttt{transformer.layers.X.[module]}) in PyTorch.

\textbf{Training Parameters.}
We use the one-cycle learning rate~\citep{smith2017cyclical} and train our models for 50 epochs using SGD optimizer. The peak learning rate for the cyclic scheduler is set to 0.1 at the 10th epoch, and the training batch size is 512. Unless specified, we add 10\% label noise to the dataset: that is, we change the label of 10\% examples to an \emph{incorrect} class chosen at random.

\section{Gradient Accounting}
\label{sec:gradient-accounting}
When training a model, at the $t^\text{th}$ epoch, $\nn_d (\theta^t) = \left(\theta_1^t,\dots,\theta_d^t\right)$ using gradient descent, the parameter update to any layer $l$ for a learning rate $\eta$ is given by:
$$ \theta_l^{t+1} = \theta_l^{t} - \eta \frac{d\mathcal{L}(\dataset,\nn_d)}{d\theta_l}$$

The quantity $\nicefrac{d\mathcal{L}(\dataset,\nn_d)}{d\theta_l}$ determines the update to the weights of the model. If knowledge were to be encoded in model weights, this quantity is an indication of the memorization potential of a layer.

\begin{figure}[t]
    \centering
    \includegraphics[width=0.75\columnwidth]{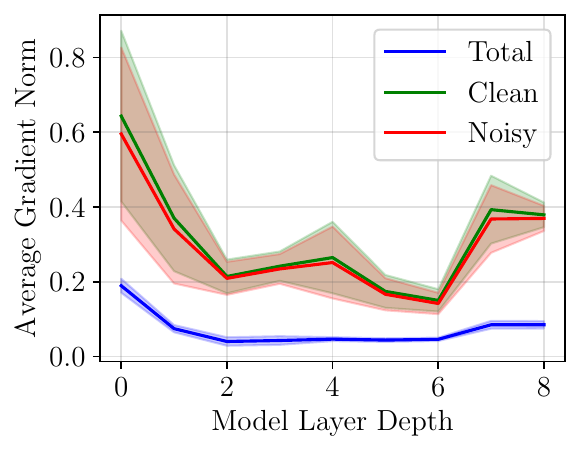}
    \caption{Gradient norm contribution from noisy examples closely follows that for clean examples even when they constitute only 10\% of the dataset. Results depicted for epochs $15$-$20$ for a ResNet-9 model trained on the CIFAR-10 dataset with 10\% label noise.}
    \label{fig:cnn_grads}
\end{figure}

\subsection{Layer-wise norm contribution}
In order to assess the contribution of gradients from clean and mislabeled examples respectively, we measure $\norm{\nicefrac{d\mathcal{L}(\dataset_c,\nn_d)}{d\theta_l}}{2}$ and $\norm{\nicefrac{d\mathcal{L}(\dataset_n,\nn_d)}{d\theta_l}}{2}$ on clean and noisy data subsets per epoch. We normalize them by the square root of the number of parameters in $\theta_l$. To combine results over multiple epochs or layers, we take the average of the individual normalized gradient norms.

\textbf{Observations and Inference.}
We observe that no layer receives significantly more contribution from memorized examples than from clean examples (Figure~\ref{fig:cnn_grads}).
However, we do notice an interesting 
the phenomenon in our results---even though the mislabeled data accounts for only 10\% of examples in the training data, their overall contribution to the gradient norms, and consequently to the model weights is the same as that of all the clean examples together. This suggests that while mislabeled examples may not be learned in particular layers of the model, they do have a large influence on all layers of the model. 
This implies that mislabeled example gradient norms are an order of magnitude higher than that for clean examples.
Results on other noise rates, datasets, and architectures can be found in Appendix~\ref{app:gradient-accounting}.

\begin{figure}[t]
\centering
\includegraphics[width=0.8\linewidth]{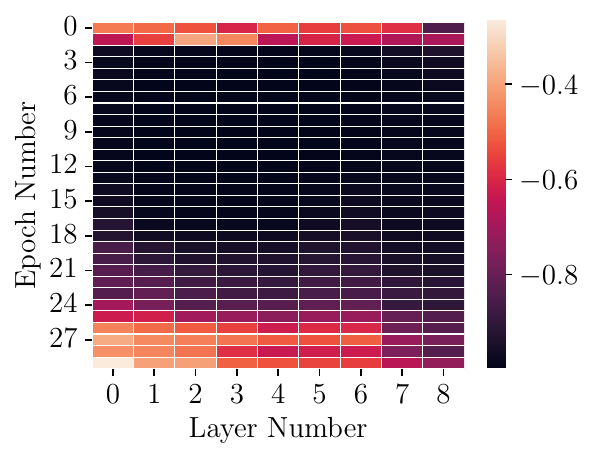}
\caption{Cosine similarity between the average gradients of clean and mislabeled examples per layer, per epoch for ResNet9 model on the CIFAR10 dataset with 10\% label noise. The memorization of mislabeled examples happens between epochs $10$--$30$ (Figure~\ref{fig:reinit}).}
\label{fig:gradient_alignment}
\end{figure}

\subsection{Clean and Noisy example gradient similarity}

Notice that the norms of total gradient contribution in Figure~\ref{fig:cnn_grads} are significantly smaller than both the gradients of clean and noisy examples respectively. This suggests that the gradient contribution from the two subsets may be canceling out each other, implying the detrimental impact of mislabeled examples on clean example training. To evaluate the same, we compute the cosine similarity between the average gradients of clean and mislabeled examples for every layer at every epoch during training.

\textbf{Observations and Inference.} We observe that the gradients of clean and mislabeled examples have an extremely negative cosine similarity. This suggests that the clean and mislabeled examples are detrimental to each other at a layer level.  However, based on the analysis of gradient similarities at a layer level, it is not possible to assert if (a) at every neuron within the layer as well, the gradients are misaligned, or (b) this is an aggregate phenomenon only occurring at the macroscopic layer level.
When contrasted with the learning curves in Figure~\ref{fig:reinit}, we note that the region of interest when the mislabeled examples are learnt belongs to epochs $10$--$30$, and the cosine similarities of the clean and mislabeled examples stay below -0.75 for all epochs and layers during the majority of this time. This indicates that overfitting to mislabeled examples may not be benign, especially when analyzed at a layer level.
Results on other noise rates, datasets, and architectures can be found in Appendix~\ref{app:gradient-accounting}.

\section{Functional Criticality of Layers}
\label{sec:functional}
\citet{zhang2019all} studied the importance of different layers of a neural network by categorizing them into robust and critical based on the robustness of the layers to re-initialization to weights from an intermediate training checkpoint. In this work, we perform this analysis in the context of memorized (mislabeled) and clean examples.
We checkpoint the model parameters $\theta^t$ for every epoch of training ($t\in [T]$). The parameters $\theta^0$ represent the model weights at initialization.

\subsection{Layer Rewinding}

\begin{figure*}[t]
    \centering
    \subcaptionbox{Training and Rewinding curves for ResNet-9 Model trained on CIFAR10 dataset\label{fig:RN9-cifar-rewind}}{
    \includegraphics[width=0.30\textwidth]{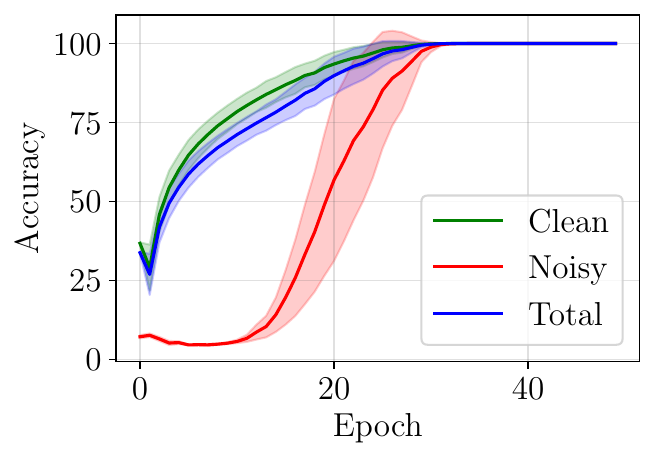}
    \includegraphics[width=0.34\textwidth]{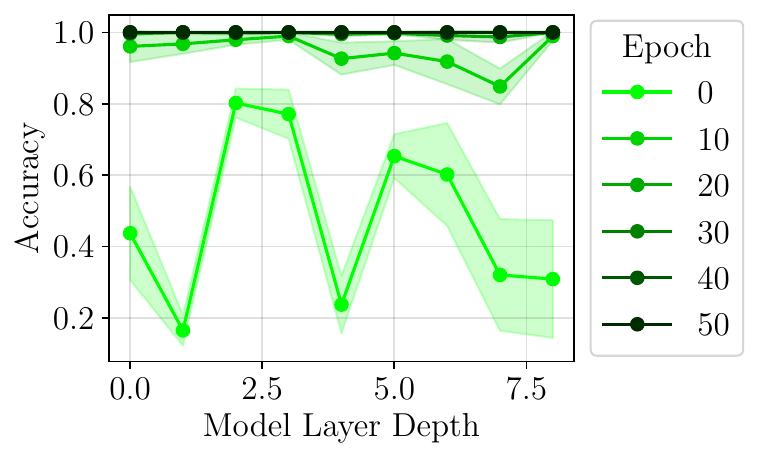}
    \includegraphics[width=0.34\textwidth]{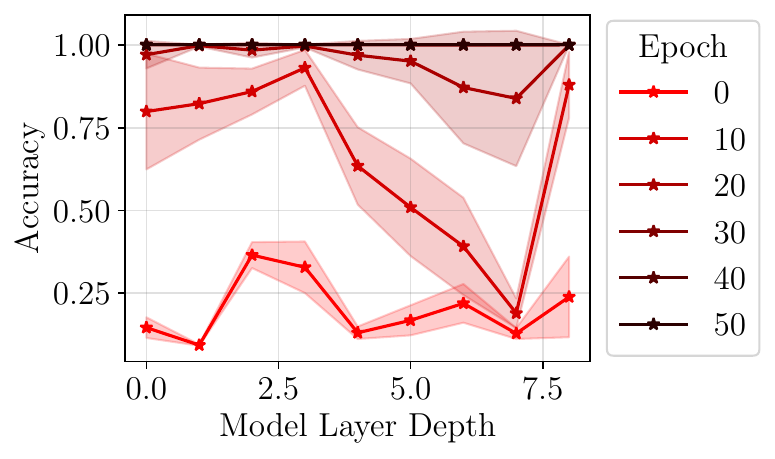}
    }\\
    \subcaptionbox{Training and Rewinding curves for ResNet-50 Model trained on CIFAR10 dataset\label{fig:RN50-cifar-rewind}}{
    \includegraphics[width=0.30\textwidth]{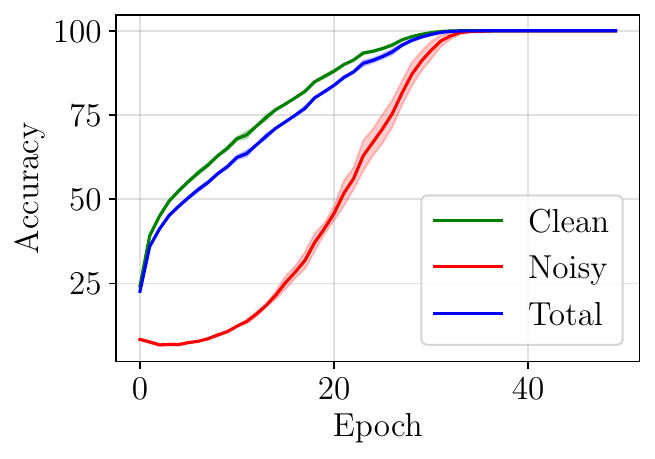}
    \includegraphics[width=0.34\textwidth]{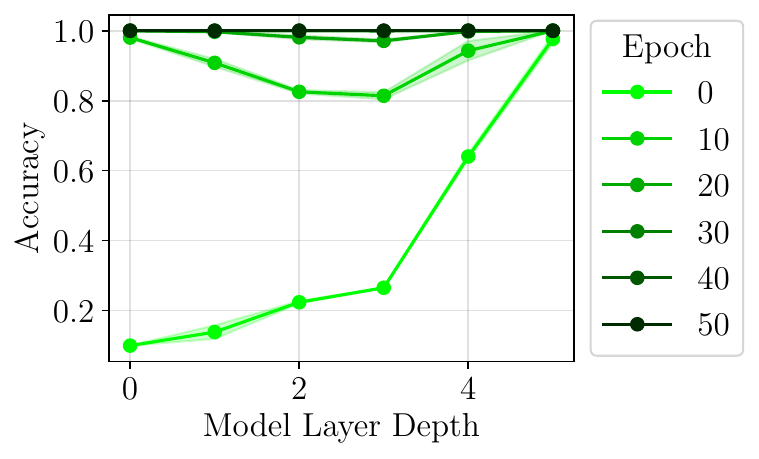}
    \includegraphics[width=0.34\textwidth]{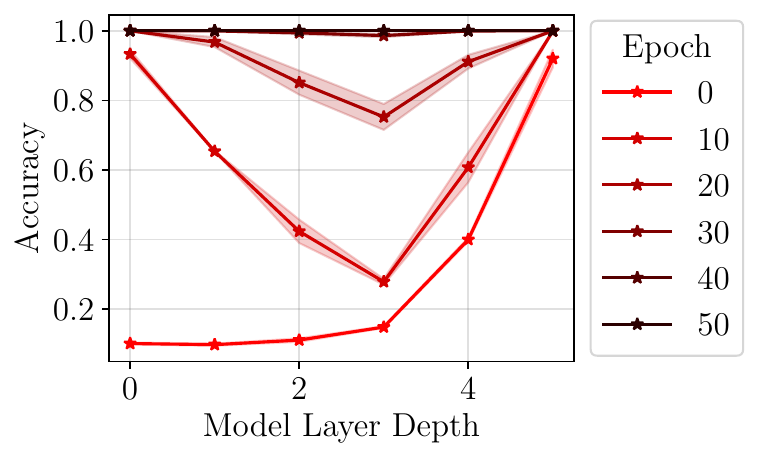}
    }\\
    \subcaptionbox{Training and Rewinding curves for ViT (small) Model trained on CIFAR10 dataset\label{fig:ViT-cifar-rewind}}{
    \includegraphics[width=0.30\textwidth]{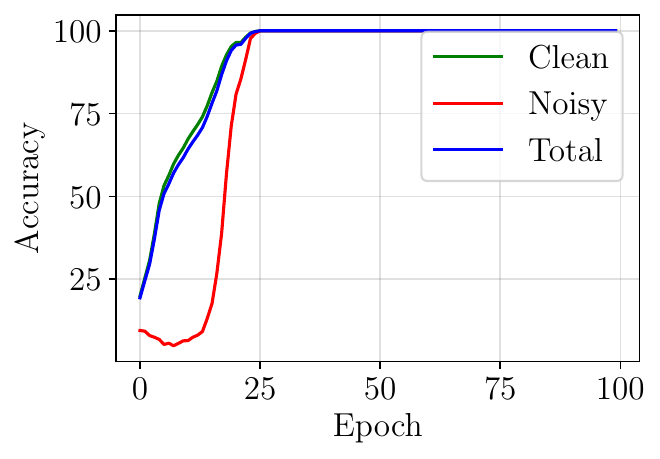}
    \includegraphics[width=0.34\textwidth]{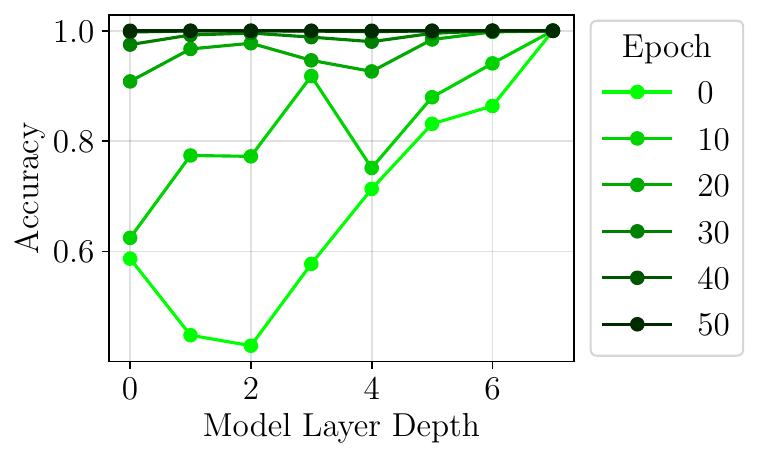}
    \includegraphics[width=0.34\textwidth]{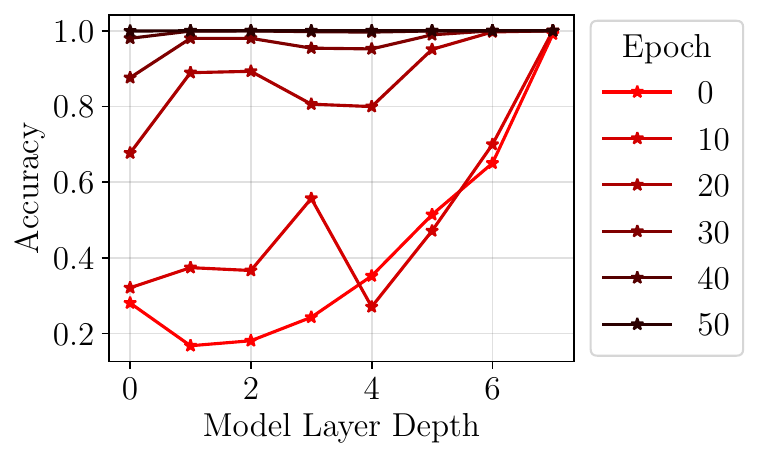}
    }
    \caption{Change in model accuracy on rewinding individual layers to a previous training epoch for clean examples (left) and mislabeled examples (right). The dataset has 10\% random label noise. Epoch 0 represents the model weights at initialization.}
    \label{fig:reinit}
\end{figure*}

For a converged model $\nn_d$ parametrized by $\left(\theta_1^T,\dots,\theta_d^T\right)$ trained for $T$ epochs,  we replace the weights of a single layer of the converged model with that of the corresponding layer in a model that was checkpointed at %
an earlier epoch during training.
We repeat the procedure for every layer (or layer group) in the model.
That is, if layer $l$ is reinitialized to epoch $t$, the new model 
$\tilde{\nn_d}$ is 
would be
parametrized by $\left(\theta_1^T,\dots,\theta_{l-1}^T,\theta_l^t,\theta_{l+1}^T\dots\theta_d^T\right)$.
Then, we evaluate the model on the training set to assess the change in training accuracy as a result of the weight rewinding.

\textbf{Observations and Inference.}
In Figure~\ref{fig:reinit} we present the accuracy upon reinitialization for clean (mid) and noisy (right) examples.
We find that rewinding the final layers to earlier epochs has low impact on the classification accuracy of the memorized examples. 
If contrasted with learning dynamics of the memorized examples (Figure~\ref{fig:reinit} (left)), we observe that the model accuracy on noisy examples stays below 20\% until the 30th epoch. Despite this, rewinding individual model layers to a checkpoint before the 30th epoch does not reduce accuracy on memorized examples. This seemingly contradicts the hypothesis that memorization happens in the last (few) layer(s). 
Secondly, on comparing the critical layers of the three architectures, we notice that the depth of critical layers for the noisy examples is different for different models. These together suggest that the location of memorization of an example  (a) is distributed across layers, (b) may depend on the exact problem instance (complexity of the model and the task), and (c) is similar to layers that are critical for clean examples as well.

\begin{figure*}[t]
    \centering
    \includegraphics[width=0.45\linewidth]{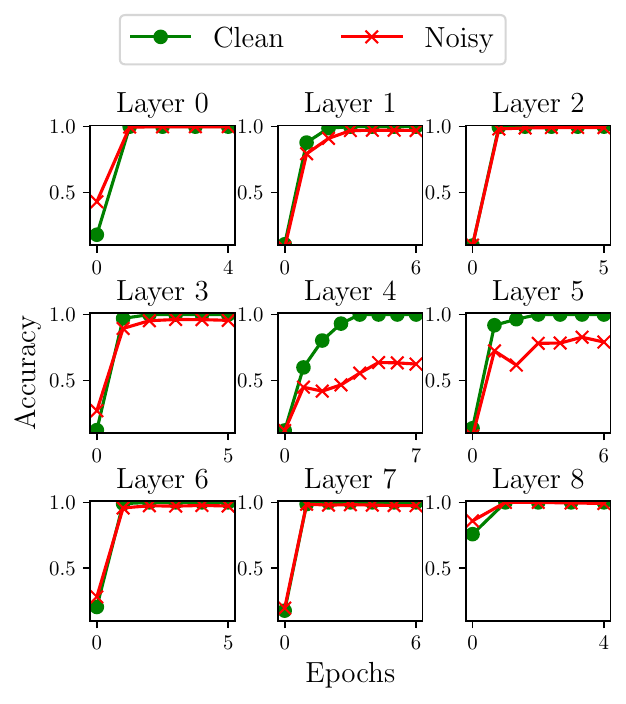}
    \hspace{10mm}\includegraphics[width=0.45\linewidth]{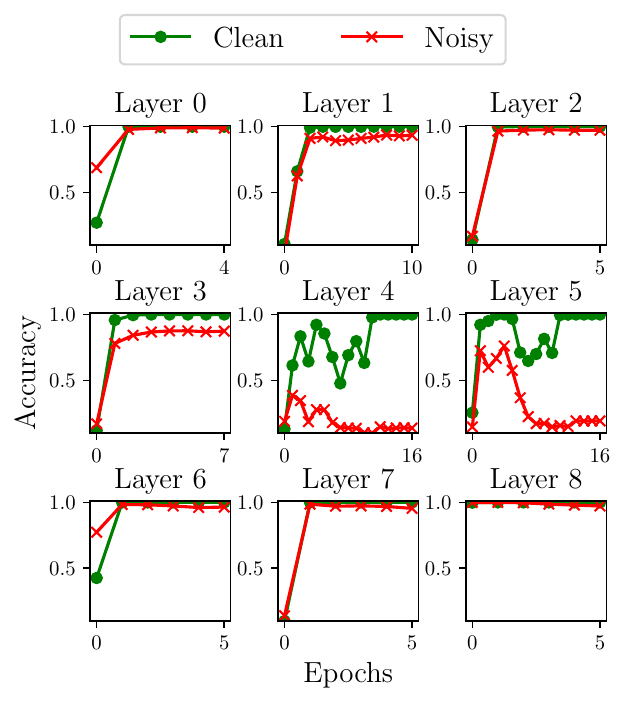}\vskip-10pt
    \caption{The impact of retraining individual layers  from scratch on only clean examples, while keeping the rest of the model frozen. The results shown are for ResNet-9 model on the CIFAR10 (left) and MNIST (right) datasets with 10\% random label noise.}
    \label{fig:retraining}
\end{figure*}

\subsection{Layer Retraining}
The experiment on layer rewinding helps us determine which layers are critical for the correct prediction of memorized examples. However, we note that the layers important for the correct prediction of mislabeled examples were also important for the prediction of clean examples. Further, the rate of learning of different layers may be an additional confounder in these results. 
To remove such a confounder, we study the impact of layer memorization by retraining individual layers after rewinding that layer back to its initialization, but only training on clean examples from there on.

When retraining layer $l$ of model $\nn_d$, $\theta_{\text{init}} = \left(\theta_1^T,\dots,\theta_{l-1}^T,\theta_l^0,\theta_{l+1}^T\dots\theta_d^T\right)$.
We then train the model for a maximum of 20 epochs using a one-cycle learning schedule with a maximum learning rate of 0.1 at the $10^{\text{th}}$ epoch~\cite{smith2017cyclical}. All other model layers are kept frozen, but the batch norm running statistics are allowed to be updated as the model trains on the clean dataset.

\textbf{Observations and Inference.}
Results of retraining individual layers show that training models on only clean examples also confers large accuracy on noisy examples (Figure~\ref{fig:retraining}). When a particular layer was reset to its initialization,   keeping the rest of the model frozen,
if the optimization objective can still achieve high accuracy on noisy examples (unseen by that layer), 
this confirms that the information required to predict correctly on memorized examples is already contained in the other layers of the model.
Therefore, the layer under question is redundant in terms of memorization. 
It is also important to note that this experiment is conclusive \emph{only} in the case when retraining on clean examples succeeds in simultaneously attaining high accuracy on mislabeled examples, making it a one-way implication that the layer was redundant for memorization.
However, if the retraining objective does not yield high accuracy on mislabeled examples, we cannot claim that it is because the layer was important for memorization, since multiple minima to the problem may exist, and the retraining objective did not reach the same minima in this instance of optimization. An example of the same can be seen in case of the MNIST dataset (Figure~\ref{fig:retraining}.b) where in Layer 5, even though the model achieves over 70\% accuracy on the mislabeled examples during the training, it finally converges at a solution that offers less than 10\% accuracy.
\section{Memorization by Individual Neurons}
\label{sec:neuron-level-memorization}
\begin{figure*}[t]
\centering
\includegraphics[width=0.33\linewidth]{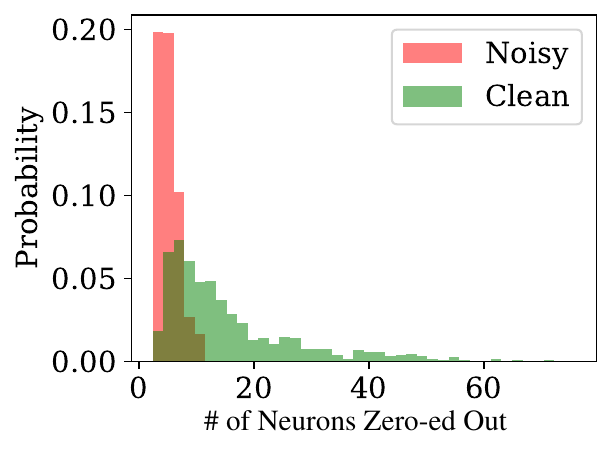}
\includegraphics[width=0.33\linewidth]{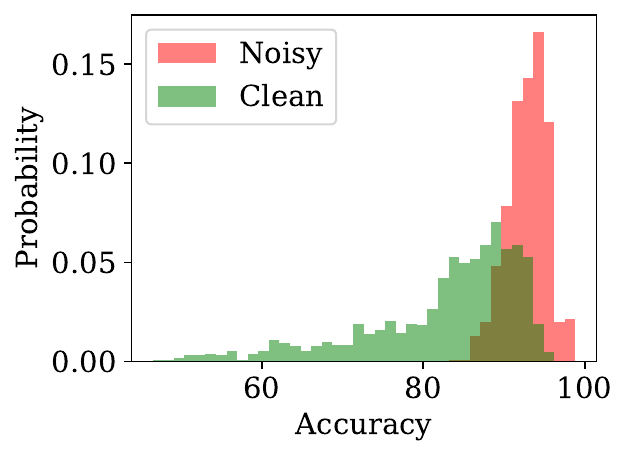}
\includegraphics[width=0.33\linewidth]{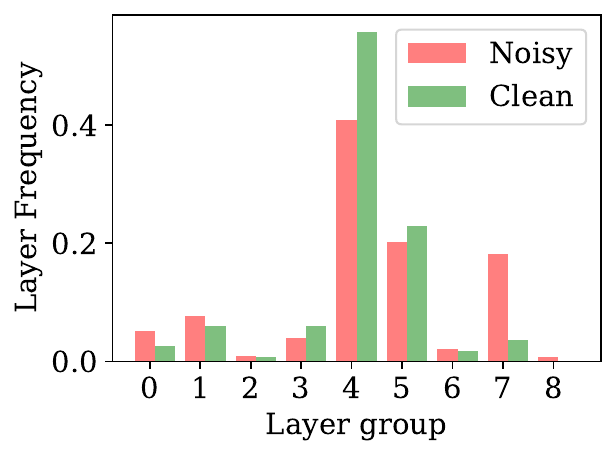}
\caption{For each example in a subset of 1000 clean and 1000 noisy examples, we iteratively remove the most important neurons from a ResNet-9 model trained on the CIFAR-10 dataset with 10\% random label noise, until the example's prediction flips. (a) Memorized examples need fewer neurons to flip their prediction. (b) Upon flipping the prediction, the drop in accuracy on the sample is much lower. (c) The most important neurons are distributed across layers in a similar way for both clean and mislabeled examples.}
\label{fig:flipping_neurons}
\end{figure*}

Experiments in the previous section highlight that it is not hard to localize the memorization of examples to particular layers of the model. This observation  points to an even more fundamental question---\emph{is the memorization of examples localized to any subset of model weights?} If this answer is in the affirmative, only then, can any attempts at localizing the memorization at a layer-level succeed.
\subsection{How many neurons does it take to predict an example?}
\label{subsec:flipping-neurons}

Through a gradient-based greedy search, we iteratively find the most important neurons for a model's prediction on a given candidate example. Then, in the case of linear layers, we zero-out a single weight in the layer. In the case of convolution layers, we zero-out a complete channel. For simplicity, we will use the term neurons to represent both scenarios. 

In order to robustly find neurons critical to the prediction of an example, we enforce the following:
(a) Preserve the prediction on the sample data by reducing the average loss on a random batch from the training set, but maximizing the loss on the input to be flipped. 
(b) Gaussian Noise is added to the input, and predictions are made based-off the average output over 5 random noise values. This is done to avoid gradient-based search to exploit the prevalence of adversarial examples for finding critical neurons. We refer to this classifier as $\overline{\nn_d}$.
Finally, for each example, we average the number of iterations required for flipping in over 30 different random training runs to make the statistic more robust.
The objective to find the most critical neuron with respect to example $(\x_i,\y_i)$ in the dataset can be written as:
\begin{align*}
        [z_l]_j^* = \arg\max_{l,j} 
        \bigg[
            \nabla_{\theta_{l}}
            \left(\ell(\overline{\nn_d}(\x_i),\y_i)
            - 
            \frac{1}{n}\mathcal{L} \left(\dataset,\overline{\nn_d}\right)
            \right)
        \bigg]_j
\end{align*}

As defined in Section~\ref{sec:prelim}, in case of convolution layers, $[z_l]_j$ represents the activations from an entire channel, whereas those from a single neuron in the case of linear layers. 
We zero-out the activation before propagating to the next layer. This procedure is repeated iteratively until the prediction on the example in question is flipped to an incorrect class.

\paragraph{Observations and Inference}

We observe that the number of neurons required to be zero-ed out in order to flip the prediction of clean examples is significantly larger than the corresponding number for noisy examples. This suggests that there is a very small number of neurons that are responsible for the correct prediction of mislabeled examples. In our experiments, for more than 90\% of memorized examples less than 10 neurons are responsible for memorization. The average of this value for the entire set of noisy and clean examples is 5.1 and 15.7 respectively (Figure~\ref{fig:flipping_neurons}.a). We also calculate the accuracy of the model on the remaining training set post-removal of neurons critical to the prediction of the example in question. The neurons corresponding to the mislabeled examples typically have a much lower impact on the population accuracy as compared to clean examples (Figure~\ref{fig:flipping_neurons}.b). The average post-removal accuracy for mislabeled examples is 10\% higher than that for clean examples.

Figure~\ref{fig:flipping_neurons}.c showcases how the critical neurons selected for both clean and mislabeled  examples most often belong to a similar set of layers, and it is not the case that important neurons for clean and mislabeled examples belong to specialized layers. This suggests that certain layers may in general be important to the model. Further, the neurons chosen are scattered over multiple layers, suggesting that it is hard to localize memorization to a particular few layers. Rather, memorization is a phenomenon confined to a select few neurons dispersed over multiple layers.

Moreover, future work may benefit by using \emph{critical neuron removal} as a metric for finding mislabeled examples from the dataset. Given any trained model, and suspect data point, find the number of neurons to be zero-ed in order to flip its prediction. If this number is lower than a pre-defined threshold, classify the example as memorized (or mislabeled). By using the same procedure, we are able to obtain an AUC of 0.89 in identifying mislabeled examples on the CIFAR10 dataset with artificially added 10\% label noise. This is competitive with existing baseline methods that achieve 0.87 (learning time~\cite{jiang2021}) and 0.93 (forgetting time~\cite{maini2022characterizing}) AUC respectively.

\subsection{Example-Tied Dropout: Directing Memorization to Specific Neurons}
\label{subsec:nested-dropout}

\begin{figure}[t]
\centering
\subcaptionbox{\label{fig:ex-drop-schema}}{  \includegraphics[width=0.57\columnwidth]{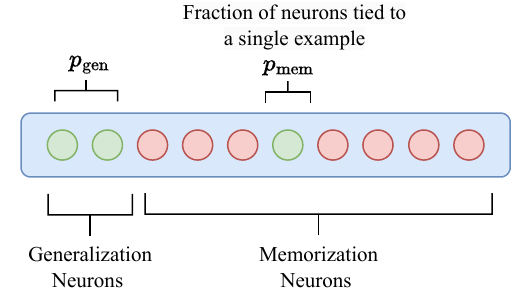}}
\subcaptionbox{\label{fig:ex-drop-prop}}{  \includegraphics[width=0.4\columnwidth]{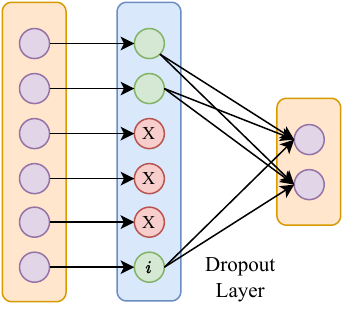}}
\caption{(a) A schematic diagram explaining the difference between the generalization and memorization neurons. At test time, we dropout all the memorization neurons. (b) Forward propagation for input tied to the $i^\text{th}$ memorization neuron}
\end{figure}

Based on our observations from Section~\ref{subsec:flipping-neurons}, we know that the information required to correctly predict on mislabeled examples is typically encoded in far fewer neurons than for clean examples. More interestingly, changing the prediction of clean examples by zero-ing out individual activations comes with a significantly higher cost on overall model accuracy on the same training set, and also on the test set. These findings motivate attempts to direct the memorization of examples to neurons that are fixed apriori.

\paragraph{Example-Tied Dropout}
To operationalize the idea of connecting a subset of neurons to every example, we assign a fixed fraction $p_\text{gen}$ of neurons per layer that are never dropped out. We call these the generalization neurons.
In the case of convolutional layers, in order to preserve the channel structure, we similarly select a fixed fraction $p_\text{gen}$ of channels that are never dropped out.
From the remaining fraction, we assign a small subset of $p_\text{mem}$ neurons per example that are activated whenever a particular example is sampled. 
The goal is to direct all the feature learning to the $p_\text{gen}$ set of neurons, and the example specific memorization neurons to the $p_\text{mem}$ set of neurons. At test time, we drop out the neurons corresponding to  $p_\text{mem}$, that is zero-out their activations, in order to adapt the model in such a way that the influence of memorization can be removed.

The experiments of example-tied dropout demonstrate that it is indeed possible to confine the memorization of examples to a pre-determined set of neurons. After each layer of a ResNet 9 model, we append a layer of `example-tied dropout' and train the model on the MNIST, SVHN, and CIFAR10 datasets in a setting with 10\% uniformly random label noise. Upon dropping the memorization neurons, the accuracy on the mislabeled examples reduced to 0.1\%, 1.4\%, and 3.1\% for the three datasets respectively, even though the model almost perfectly fits on all the noisy data at training time.  In all cases, the effective impact on the clean examples is only 0.8\%, 4.2\%, and 9.2\% respectively, also lowering the generalization gap between train and test performance. This suggests that the $p_\text{gen}$ fraction of (generalization) neurons already contained the features required to classify the prototypical examples of the dataset. 

\begin{table}[t]
\caption{Dropping out memorization neurons leads to a sharp drop in accuracy on mislabeled examples with a minor impact on prediction on clean and unseen examples.}
\scalebox{0.88}{
\begin{tabular}{@{}lrrrrrr@{}}
\toprule
\multirow{2.5}{*}{Dataset} & \multicolumn{3}{c}{Before Dropout}                                                  & \multicolumn{3}{c}{After Dropout}                                                \\ \cmidrule(l){2-7} 
                         & \multicolumn{1}{l}{Clean} & \multicolumn{1}{l}{Noisy} & \multicolumn{1}{l}{Test}   & \multicolumn{1}{l}{Clean} & \multicolumn{1}{l}{Noisy} & \multicolumn{1}{l}{Test} \\ \midrule
CIFAR10                  & 99.9\%                     & 99.3\%                    & \multicolumn{1}{r|}{79.3\%} & 90.8\%                    & 3.1\%                     & 82.7\%                   \\
MNIST                    & 100\%                     & 100\%                     & \multicolumn{1}{r|}{99.0\%} & 99.2\%                    & 0.1\%                     & 99.3\%                   \\
SVHN                     & 99.9\%                     & 99.6\%                     & \multicolumn{1}{r|}{89.5\%} & 95.8\%                    & 1.4\%                     & 89.6\%                   \\ \bottomrule
\end{tabular}
}
\label{tab:ex-tied-dropout}
\end{table}

\begin{figure}[t]
  \floatbox[{\capbeside\thisfloatsetup{capbesideposition={right,top},capbesidewidth=0.4\textwidth}}]{figure}[\FBwidth]
  {\caption{Most of the clean examples that are forgotten when dropping out the neurons responsible for memorization in the case of Example-tied dropout were either mislabeled or inherently ambiguous and unique requiring memorization for correct classification.}\label{fig:et-mnist-clean-errors}}
  {\includegraphics[width=0.5\textwidth]{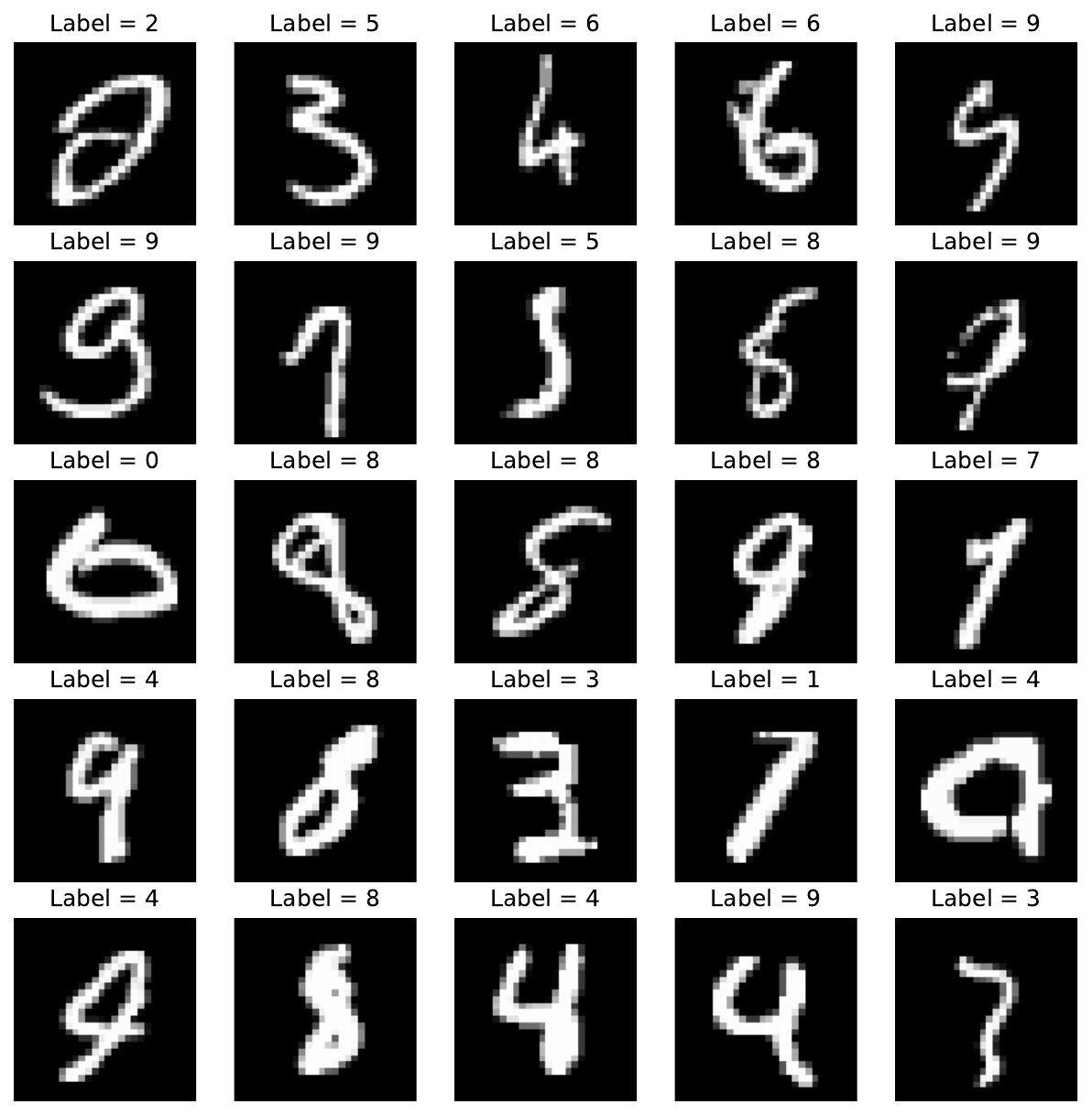}}
  \end{figure}

\paragraph{Analysis of clean examples forgotten}
To understand the reason for the drop in accuracy on clean examples on removing the memorization neurons, we find that most of the examples that the model no longer correctly predicts are in fact either mislabeled and ambiguous or exceptional and unique examples from the dataset that the model must memorize in order to correctly predict on (Figure~\ref{fig:et-mnist-clean-errors}).

\begin{table*}[t]
\caption{Impact on final \textcolor{OliveGreen}{clean} and \textcolor{Mahogany}{noisy} example accuracy (training set) when changing the values of $p_{\text{mem}}$ and $p_{\text{gen}}$.
Each cell represents the \textcolor{OliveGreen}{clean}, \textcolor{Mahogany}{noisy} example accuracy after dropping out all the memorization neurons during test time.
}
\centering
\begin{tabular}{ccccccc}
\toprule
\multicolumn{1}{c}{$p_{\text{mem}} / p_{gen}$} & \multicolumn{1}{c}{0.1} & \multicolumn{1}{c}{0.2} & \multicolumn{1}{c}{0.3} & \multicolumn{1}{c}{0.4} & \multicolumn{1}{c}{0.5} & \multicolumn{1}{c}{0.6} \\
\midrule
0.1 & \textcolor{OliveGreen}{84.2\%}, \textcolor{Mahogany}{2.9\%} & \textcolor{OliveGreen}{86.9\%}, \textcolor{Mahogany}{2.8\%} & \textcolor{OliveGreen}{87.1\%}, \textcolor{Mahogany}{2.9\%} & \textcolor{OliveGreen}{89.9\%}, \textcolor{Mahogany}{3.1\%} & \textcolor{OliveGreen}{92.0\%}, \textcolor{Mahogany}{3.9\%} & \textcolor{OliveGreen}{93.0\%}, \textcolor{Mahogany}{4.9\%} \\
0.2 & \textcolor{OliveGreen}{86.9\%}, \textcolor{Mahogany}{2.8\%} & \textcolor{OliveGreen}{88.2\%}, \textcolor{Mahogany}{3.0\%} & \textcolor{OliveGreen}{90.8\%}, \textcolor{Mahogany}{3.1\%} & \textcolor{OliveGreen}{91.7\%}, \textcolor{Mahogany}{3.4\%} & \textcolor{OliveGreen}{93.6\%}, \textcolor{Mahogany}{4.7\%} & \textcolor{OliveGreen}{95.0\%}, \textcolor{Mahogany}{6.9\%} \\
\bottomrule
\end{tabular}
\label{tab:hp-change}
\end{table*}

\begin{table}[t]
  \centering
  \caption{Comparison of Accuracy on \textcolor{OliveGreen}{clean}, \textcolor{Mahogany}{noisy} training examples after training with various methods that activate only a small fraction of the network during training time.}
  \label{tab:accuracy_comparison}
  \begin{tabular}{lc}
    \toprule
    {Method} & {Accuracy} \\
    \midrule
    Standard Dropout (p = 0.4) & \textcolor{OliveGreen}{100\%,} \textcolor{Mahogany}{99.5\%} \\
    Sparse Network (s = 0.4) & \textcolor{OliveGreen}{100\%,} \textcolor{Mahogany}{99.8\%} \\
    Example-Tied ($p_{\text{gen}}$ = 0.4, $p_{\text{mem}}$ = 0.2) & \textcolor{OliveGreen}{90.8\%,} \textcolor{Mahogany}{3.10\%} \\
    \bottomrule
  \end{tabular}
\end{table}

\paragraph{Impact of Hyperparameter Tuning}
To assess the impact of the choice of values of $p_{\text{gen}}$ and $p_{\text{mem}}$ on the ability of example-tied dropout to localize memorization in neurons, we run a grid search on 12 different parameter combinations. 
Results for the change in the efficacy of the method with a change in these values are presented in Table~\ref{tab:hp-change}. We find that the method is robust to a large range of values (and combinations) of $p_{\text{gen}}$ and $p_{\text{mem}}$. In general, we observe the trend that as we increase the capacity of the generalization neurons, the localization of memorization in the memorization neurons keeps decreasing (that is, the noisy example accuracy upon dropping out the memorization neurons increases). At the same time, this also results in an increase in clean example accuracy after dropping out of memorization neurons.

\paragraph{Benchmarking with other baselines}
 For a given value of $p_\text{gen}$, we select two equivalent models that (a) use standard dropout with p = $p_\text{gen}$, and (b) use a sparse network with remaining parameter fraction = $p_\text{gen}$. Then, we finally compare the accuracy of the model on clean and noisy training examples after training for 50 epochs on the CIFAR-10 dataset, using the ResNet-9 model. In the case of example-tied dropout, we drop the memorization neurons while measuring the accuracy in Table~\ref{tab:accuracy_comparison}. We see that at the same sparsity level when example-tied dropout is able to direct memorization to a chosen set of neurons, sparse and standard dropout methods are still capable of overfitting to the training set (including mislabeled examples).

 \section{Memorization of Atypical Examples}
 The experiments until now  operationalize memorization through `mislabeled examples'. Since the label for these examples does not obey the class semantics, it is convenient to categorize any `correctly' predicted mislabeled example as `memorized'. However, this is a very strong notion of memorization and prior work has observed the existence of `atypical examples' that also end up getting memorized~\cite{feldman2020neural}. Such a definition of memorization utilizes the `leave-one-out' form of memorization. An example is said to be memorized if its prediction changes significantly when it is present versus absent in the training set of the model. \citet{jiang2021} approximated a consistency score (C-score) for each example by bootstrapping the leave-one-out memorization.
 In particular, a low C-score (in the [0,1] range) implies that the model will need to memorize an example. 
 We use a threshold of 0.5 and select all examples below the threshold to construct the `memorized set'. This set is of the size $\sim5000, ~\sim1000$  examples for the MNIST and CIFAR10 datasets respectively. We do not add any label noise in this set of experiments.
 
\begin{figure}
    \centering
    \includegraphics[width=\linewidth]{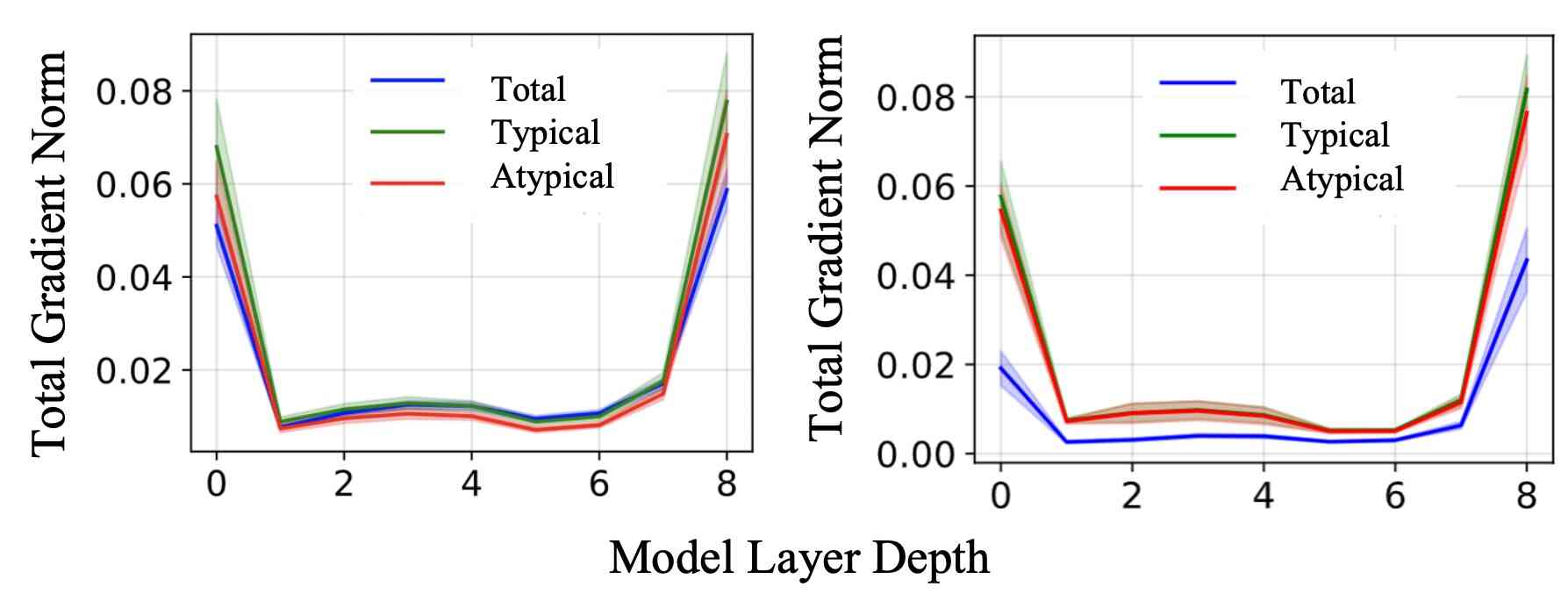}
    \caption{Gradient Accounting for Atypical Examples when training a ResNet-9 model on CIFAR-10 (left) and MNIST (right).}
    \label{fig:grad_acc_cscore}\vspace{-0.07\textwidth}
 \end{figure}

 \textbf{Gradient Accounting.}
 We find that the results are consistent with past results that study the memorization of mislabeled examples. In particular, the aggregate gradient contribution from atypical examples is similar to typical examples despite their significantly smaller size (Figure~\ref{fig:grad_acc_cscore}).

\begin{figure}
      \centering
      \includegraphics[width=\linewidth]{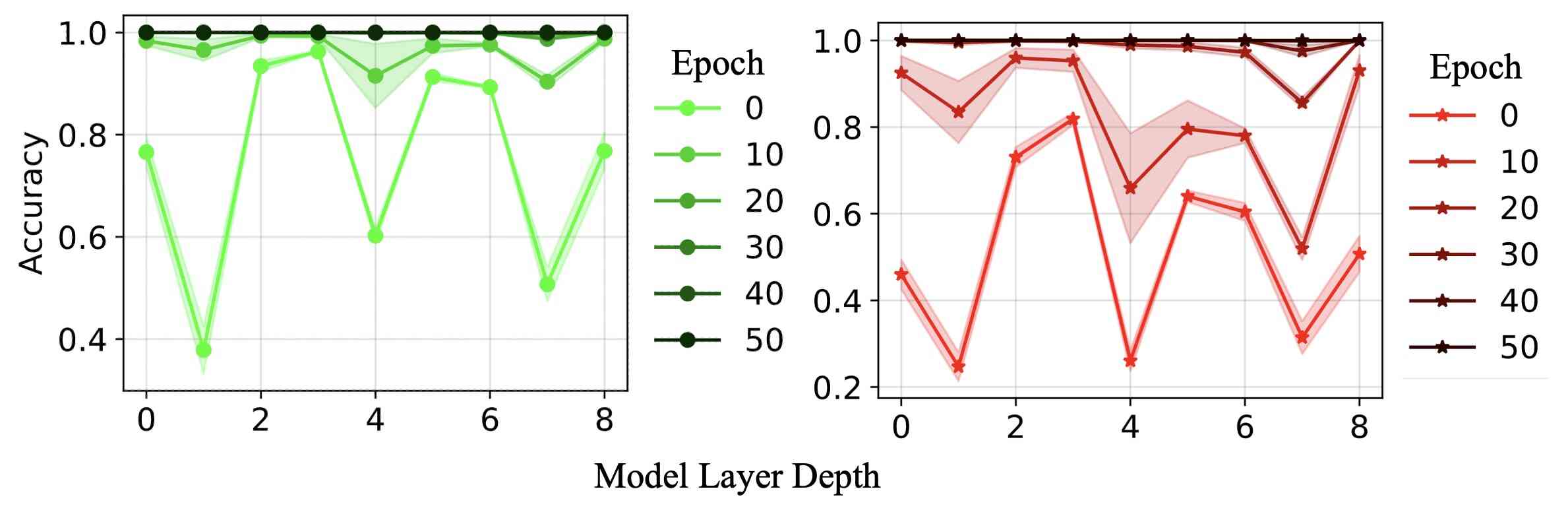}
      \caption{Layer Rewinding for Atypical Examples in CIFAR-10}
      \label{fig:rewind_cscore}\vspace{-0.07\textwidth}
  \end{figure}

 \textbf{Layer Rewinding.}
  Rewinding plots in  Figure~\ref{fig:rewind_cscore} suggest that memorization of atypical examples is dispersed across various layers of the model, which is  also consistent with the results for mislabeled examples. 

 \textbf{Layer Retraining.}
We find that in the retraining experiment, all layers can be retrained by seeing only examples with C-score $>0.5$ such that the “memorized” examples also achieve ~100\% accuracy. This suggests how the localization of memorization is even more challenging in such OOD/atypical example settings (Appendix~\ref{app:subsubsec:retraining},  Figure~\ref{app:fig:retraining-atypical}).

\section{Discussion}

In this work, we perform a systematic study to identify if the memorization of examples is confined to particular locations within a network. Contrary to previous hypotheses, our results show that memorization is often dispersed across multiple layers.
In particular, our results have important implications for both the privacy and the generalization of neural networks. From a privacy point of view, unlearning specific layers \textit{may not} help the model forget memorized examples; and for generalization, rewinding or strongly regularizing only a small set of layers may be sub-optimal since memorization can not be localized to a few layers.

On the other hand, our work showed that memorization is localized to a few neurons dispersed across different model layers. We introduce a novel mechanism to direct memorization to a predetermined subset of neurons,  using example-tied dropout. At test time, practitioners may throw away such `memorized' neurons. This has positive implications in the field of machine unlearning and training with noisy data.

\section*{Acknowledgements}
We gratefully acknowledge the NSF (FAI 2040929 and IIS2211955), UPMC, Highmark Health, Abridge, Ford Research, Mozilla, the PwC Center, Amazon AI, JP Morgan Chase, the Block Center, the Center for Machine Learning and Health, and the CMU Software Engineering Institute (SEI) via Department of Defense contract FA8702-15-D-0002, for their generous support of ACMI Lab’s research. Pratyush Maini is supported by funding from the DARPA GARD program.
\bibliography{paper}
\bibliographystyle{paper}

\clearpage

\appendix

\onecolumn
\section*{\centering Appendix: Can Neural Network Memorization Be Localized? \\ \textcolor{white}{.}}

\section{Gradient Accounting}
\label{app:gradient-accounting}
\subsection{Progression of gradient contribution through training}
\begin{figure*}[h]
    \centering
    \subcaptionbox{Gradient norm contribution from clean and noisy examples over the course of epochs 0-5, 5-10, and 10-15 respectively.\label{}}{
    \includegraphics[width=0.33\textwidth]{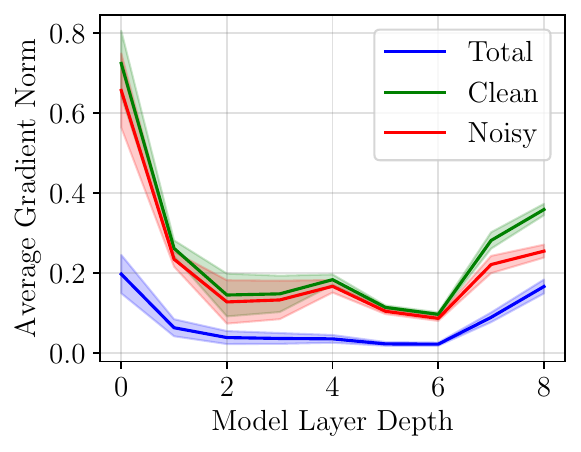}
    \includegraphics[width=0.33\textwidth]{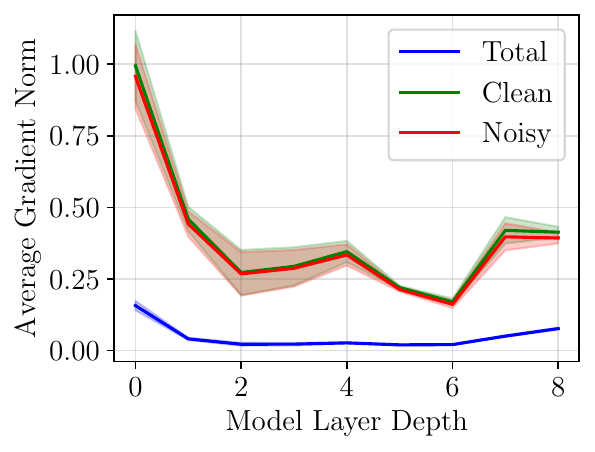}
    \includegraphics[width=0.33\textwidth]{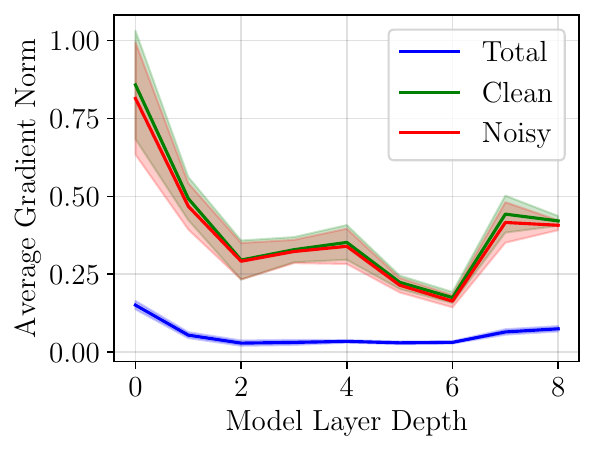}
    }\\
    \subcaptionbox{Gradient norm contribution from clean and noisy examples over the course of epochs 15-20, 20-25, and 25-30 respectively.\label{}}{
    \includegraphics[width=0.33\textwidth]{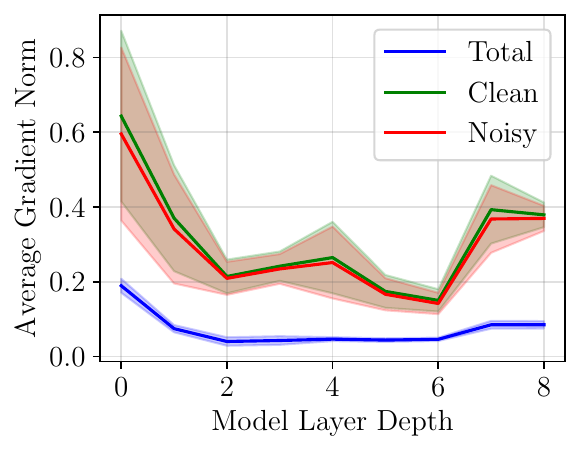}
    \includegraphics[width=0.33\textwidth]{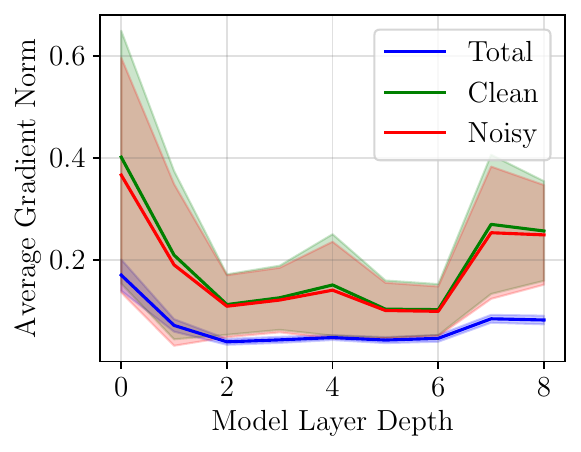}
    \includegraphics[width=0.33\textwidth]{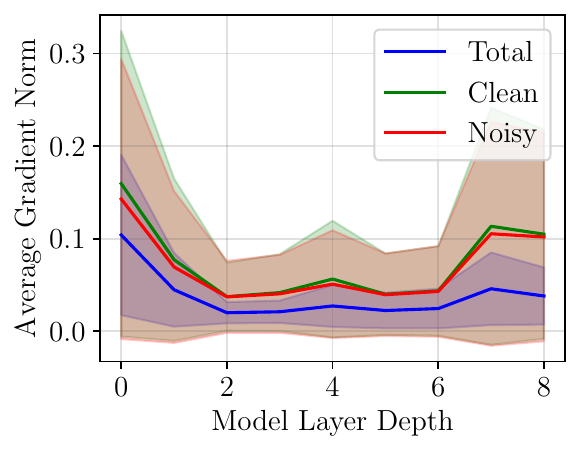}
    }\\
    \subcaptionbox{Gradient norm contribution from clean and noisy examples over the course of epochs 30-35, 35-40, and 40-45 respectively.\label{}}{
    \includegraphics[width=0.33\textwidth]{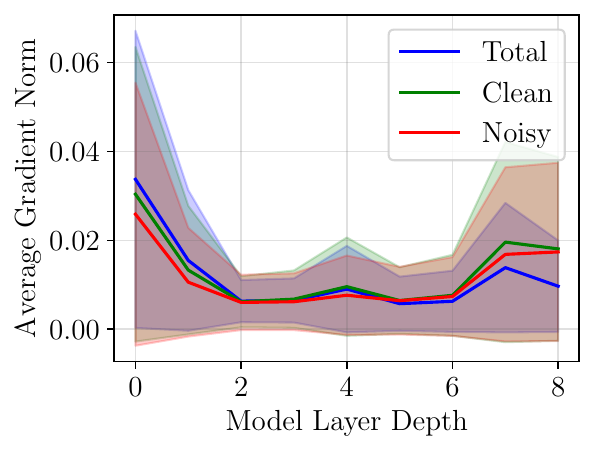}
    \includegraphics[width=0.33\textwidth]{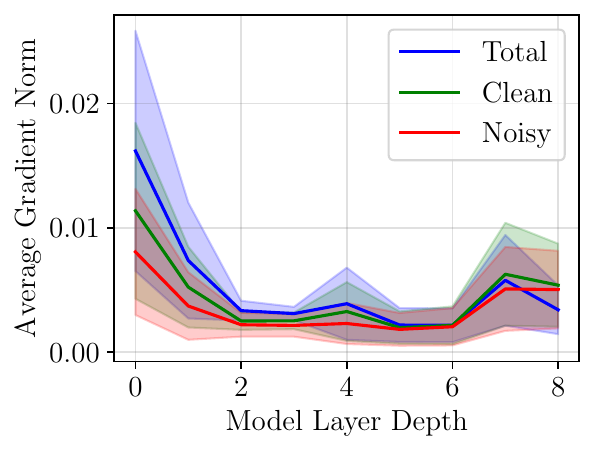}
    \includegraphics[width=0.33\textwidth]{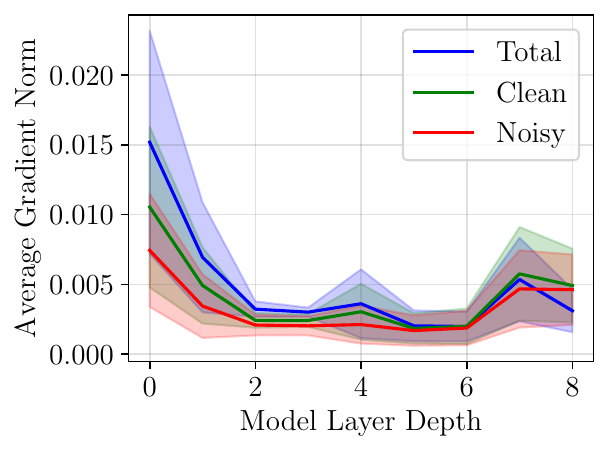}
    }
    \caption{Total gradient norm contribution (averaged over different epochs)  from mislabeled and clean examples over the course of training a ResNet-9 model on the CIFAR-10 dataset with 10\% label noise.}
    \label{fig:grads-contribution}
\end{figure*}

In this section, we provide fine-grained results showing the variation in gradient norms of clean and noisy examples over the course of training. Recall from the learning curves in the main paper that the learning of mislabeled examples primarily takes place in the epochs 10-30.
When we look at the gradient norms of clean and noisy examples over the course of training, we see that the gradient norms contributed by just 10\% of noisy examples is nearly as large as the gradient norms contributed by all the clean examples. This is true for all the epochs in the range 0-50. This is shown in Figure \ref{fig:grads-contribution}. This suggests that even though the noisy examples influence model weights throughout training, and not just in the final few epochs.

\subsection{Fine-Grained Analysis of Gradient Contribution}
\label{app:subsubsec:radient-progression-with-noise}
\begin{figure*}[t]
\begin{center}

\subcaptionbox{Noise Rate = 1\%\label{}}{
\includegraphics[height=0.25\linewidth]{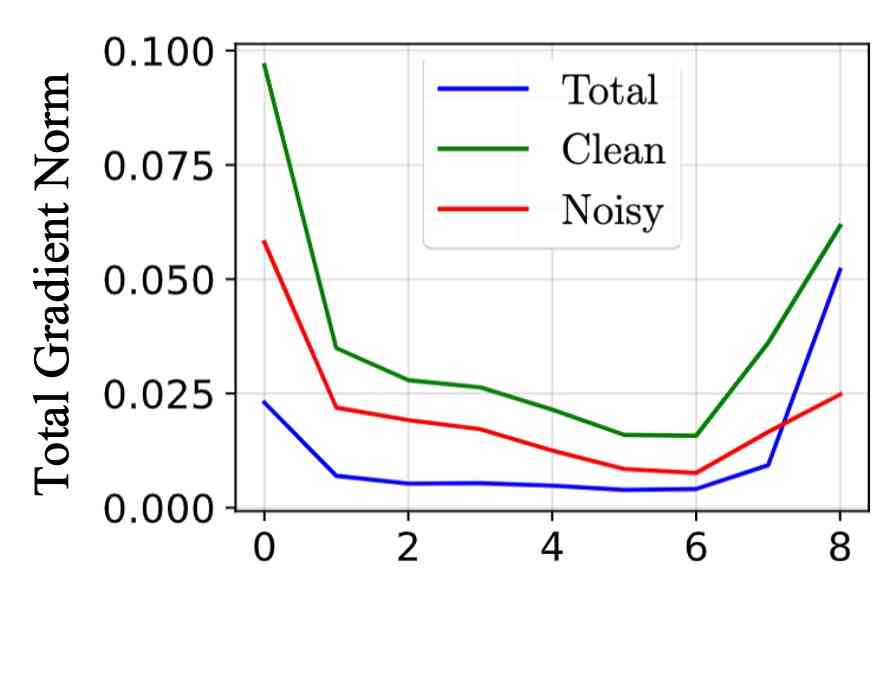}}
\subcaptionbox{Noise Rate = 2\%\label{}}{
\includegraphics[height=0.25\linewidth]{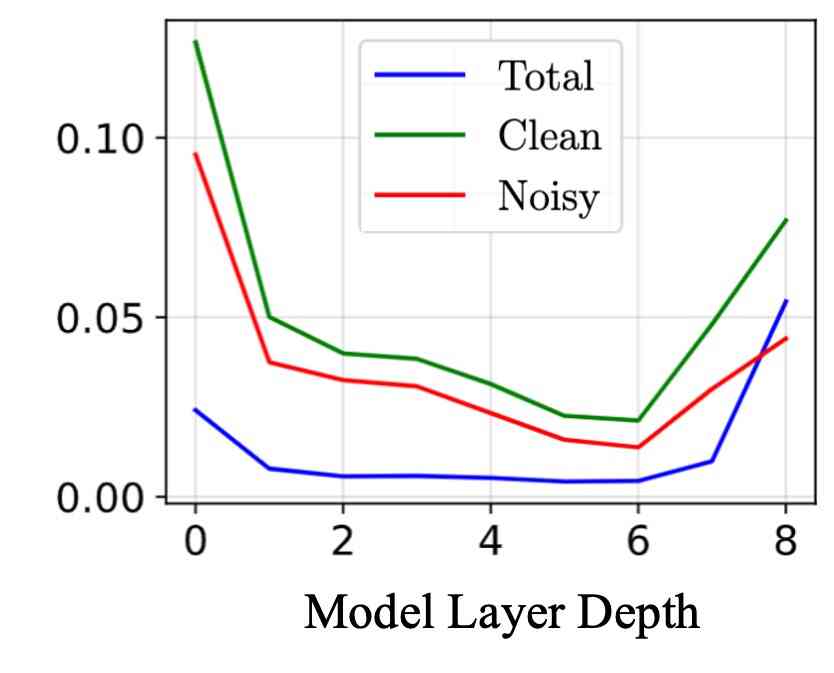}}
\subcaptionbox{Noise Rate = 5\%}{
\includegraphics[height=0.25\linewidth]{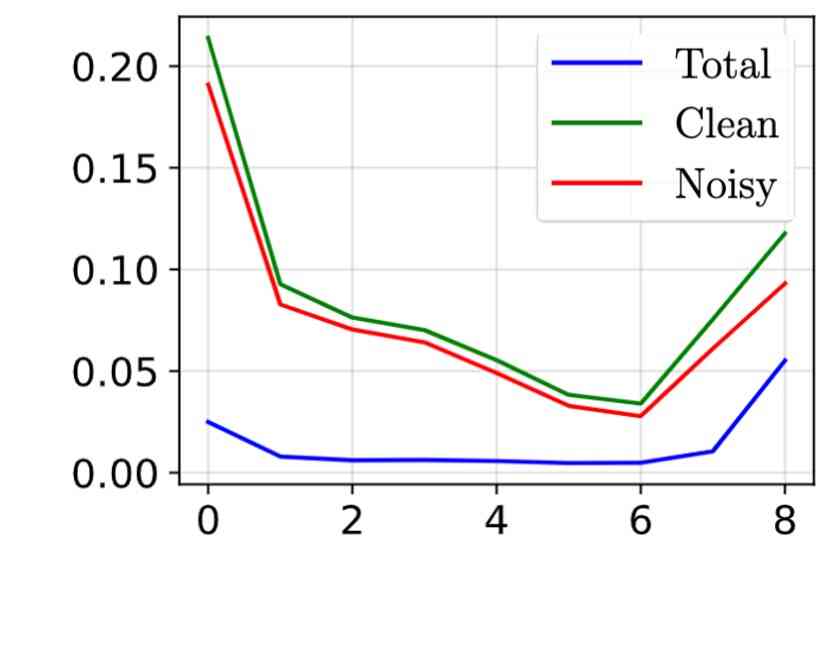}}
\\
\subcaptionbox{Noise Rate = 10\%\label{}}{
  \includegraphics[height=0.25\linewidth]{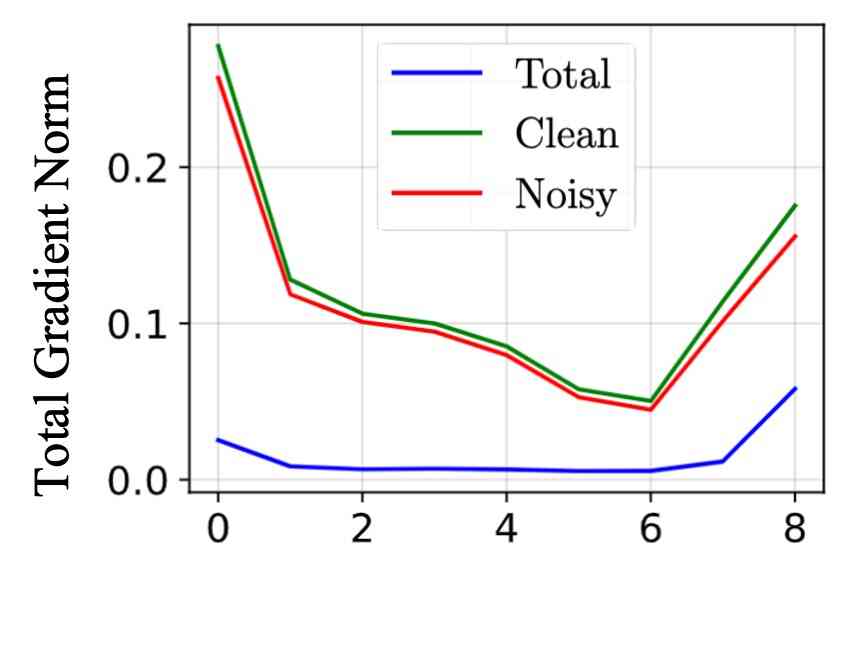}}
\subcaptionbox{Noise Rate = 20\%\label{}}{
  \includegraphics[height=0.25\linewidth]{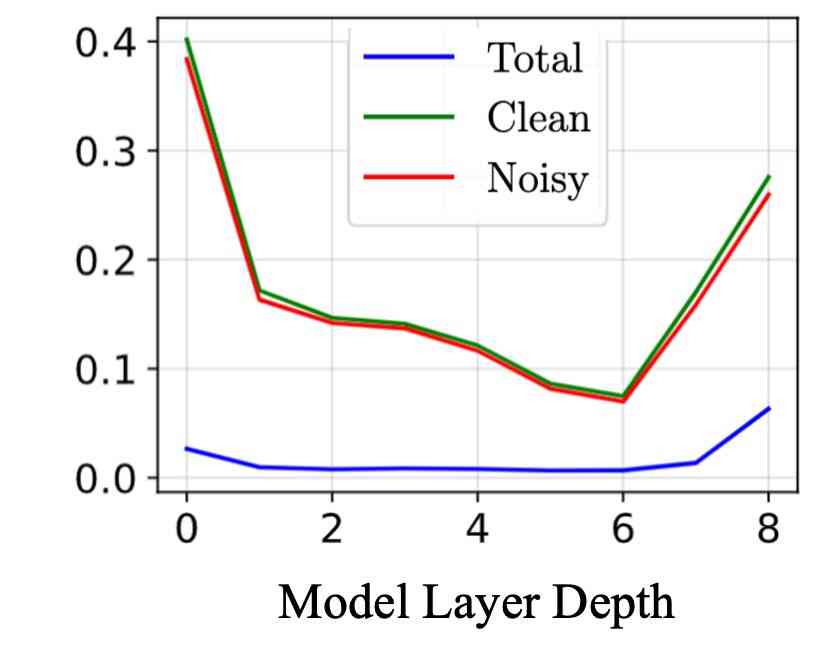}}
\subcaptionbox{Noise Rate = 50\%}{
  \includegraphics[height=0.25\linewidth]{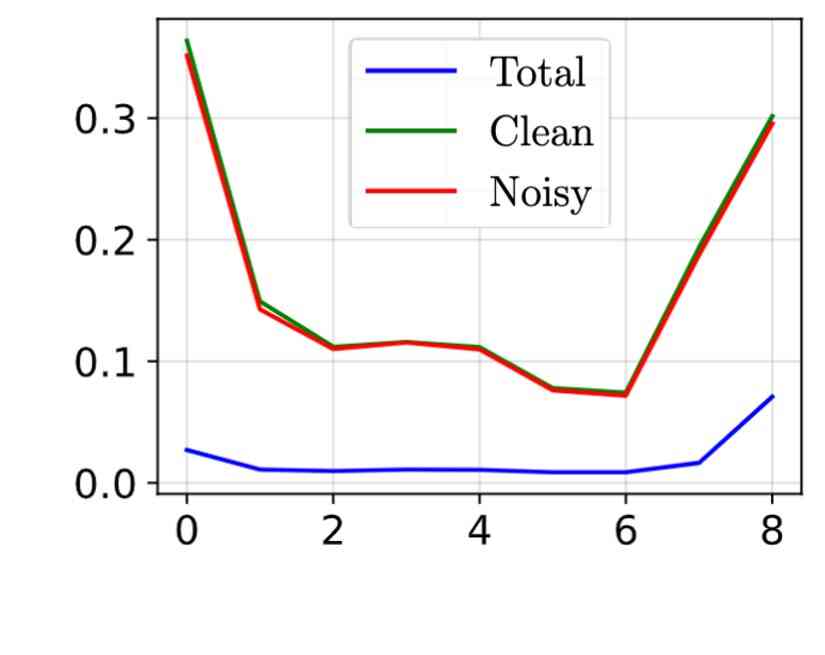}}
\\
\subcaptionbox{Noise Rate = 70\%\label{}}{
  \includegraphics[height=0.25\linewidth]{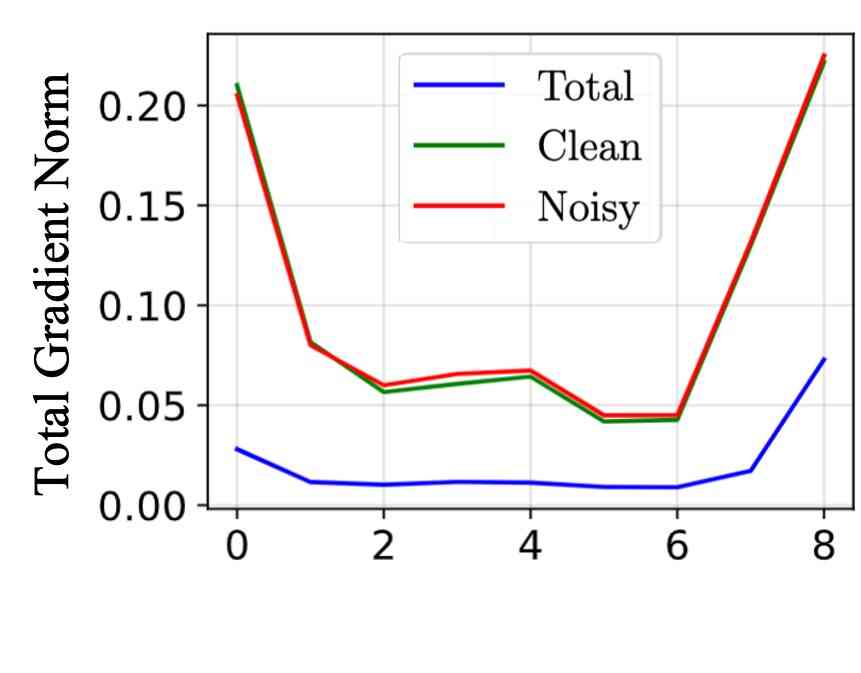}}
\subcaptionbox{Noise Rate = 90\%\label{}}{
  \includegraphics[height=0.25\linewidth]{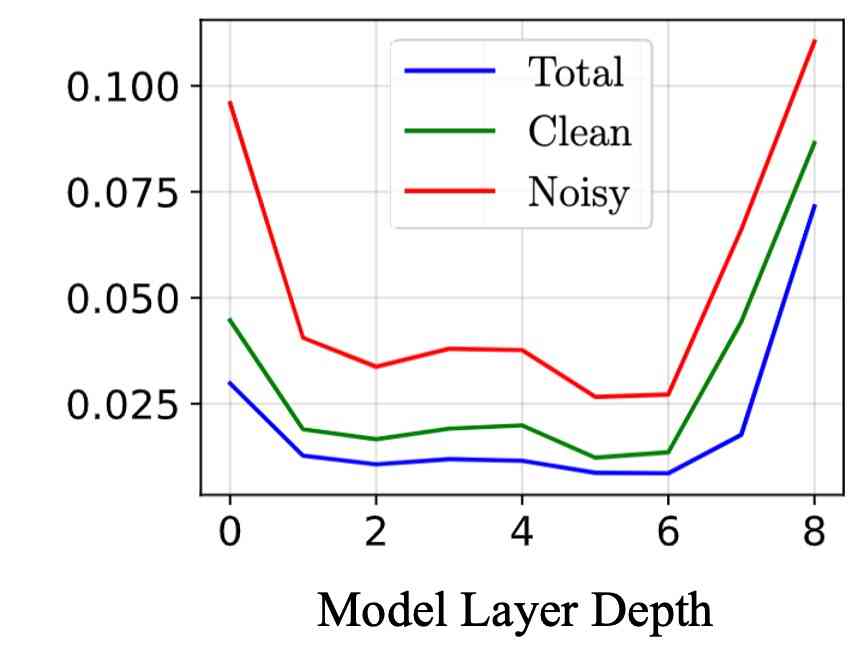}}
\subcaptionbox{Noise Rate = 95\%}{
  \includegraphics[height=0.25\linewidth]{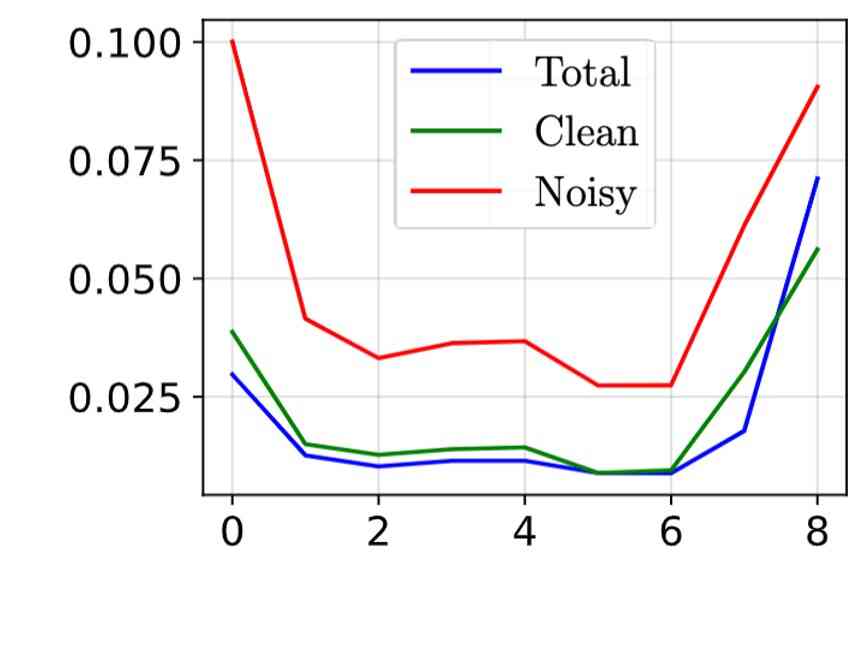}}
  \\
  \subcaptionbox{Noise Rate = 98\%\label{}}{
  \includegraphics[height=0.25\linewidth]{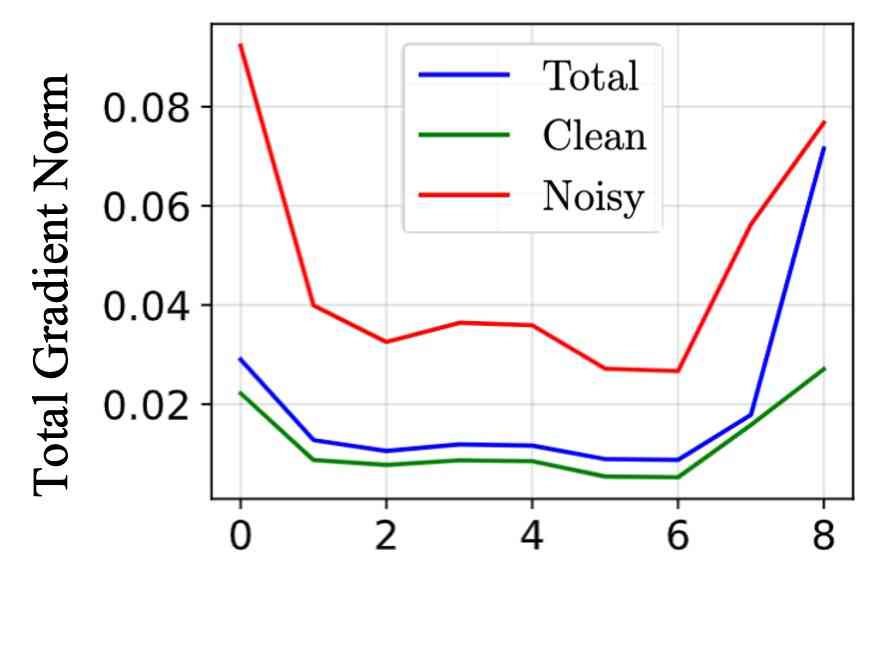}}
\subcaptionbox{Noise Rate = 99\%\label{}}{
  \includegraphics[height=0.25\linewidth]{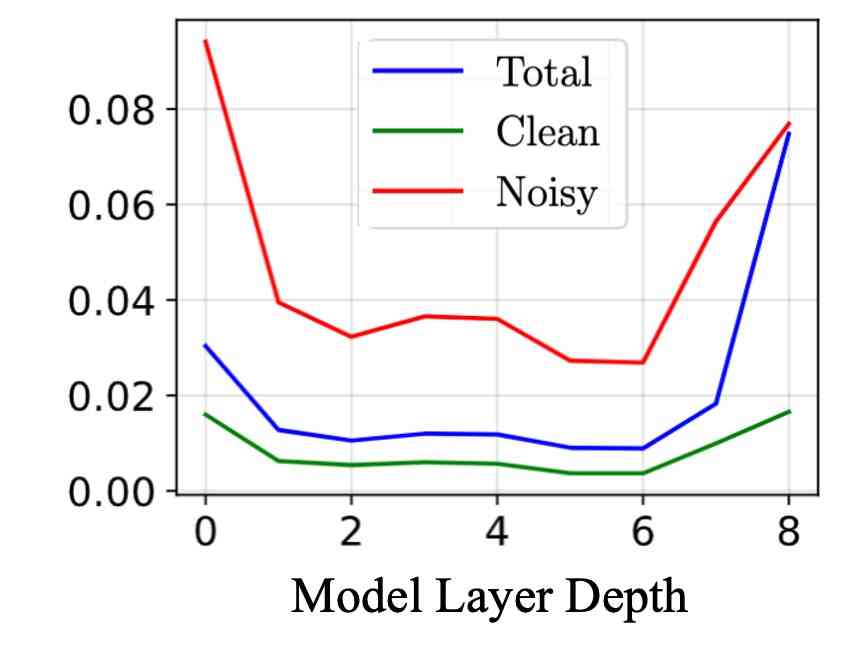}}
\hspace{0.33\textwidth}
\caption{We trained ResNet-9 models at varying percentages of random noise in the CIFAR-10 dataset. The graphs show the gradient norm contribution (averaged across 50 epochs of training) of clean and noisy examples. Detailed inference is discussed in~\S~\ref{app:subsubsec:radient-progression-with-noise}}
 \label{app:fig:gradient-progression-with-noise}
 \end{center}
\end{figure*}

Recall that we only average the gradient norms across all epochs, and do not normalize them by the number of examples.
 That is, if we were to sum up the clean and noisy example gradients, then it would yield the total gradients. 
In our gradient contribution analysis, we fond the surprising trend that the overall gradient contribution from clean and noisy examples is nearly identical even though their proportions are vastly different. 
 To understand this trend better,
 we extended the evaluation at varying noise rates from 1\% all the way up to 99\% and present the graphs in Figure~\ref{app:fig:gradient-progression-with-noise}. All results are presented on the CIFAR-10 dataset by training ResNet-9 models for 50 epochs at varying percentages of random noise in the dataset.

We observe that the contribution from mislabeled examples goes from being only slightly lower than clean examples, at 1\% noise rate, to very high at 99\% noise rate. One explanation for the high overall gradient contribution from noisy examples at very low noise rates (such as 1\%) is that clean examples learn via coherent gradients~\cite{Chatterjee2020Coherent} whereas mislabeled examples have to individually contribute significantly more in order to be memorized (or predicted correctly). This explains that mislabeled (and memorized) examples have a significant contribution to the model weights when compared to clean examples, and surprisingly, this phenomenon holds at even very low noise rates (such as 1\%).

\subsection{Gradient Analysis for different models and noise rates}
\label{app:subsubsec:gradients-all}
We supplement the results of conclusions of the main paper regarding the dispersed contribution of noisy example gradients by repeating our experiments at 3 different noise rates (5\%, 10\%, 20\%), 3 models (ResNet-9, ResNet-50, ViT-small), and 3 different datasets (CIFAR-10, MNIST, SVHN). Each experiment is averaged over 3 random seeds (except those for ViT).

\begin{figure*}[h]
    \centering
    \subcaptionbox{Gradient norm contribution from clean and noisy examples on when label noise is 5\% in the CIFAR-10 dataset.\label{}}{
    \includegraphics[width=0.9\textwidth]{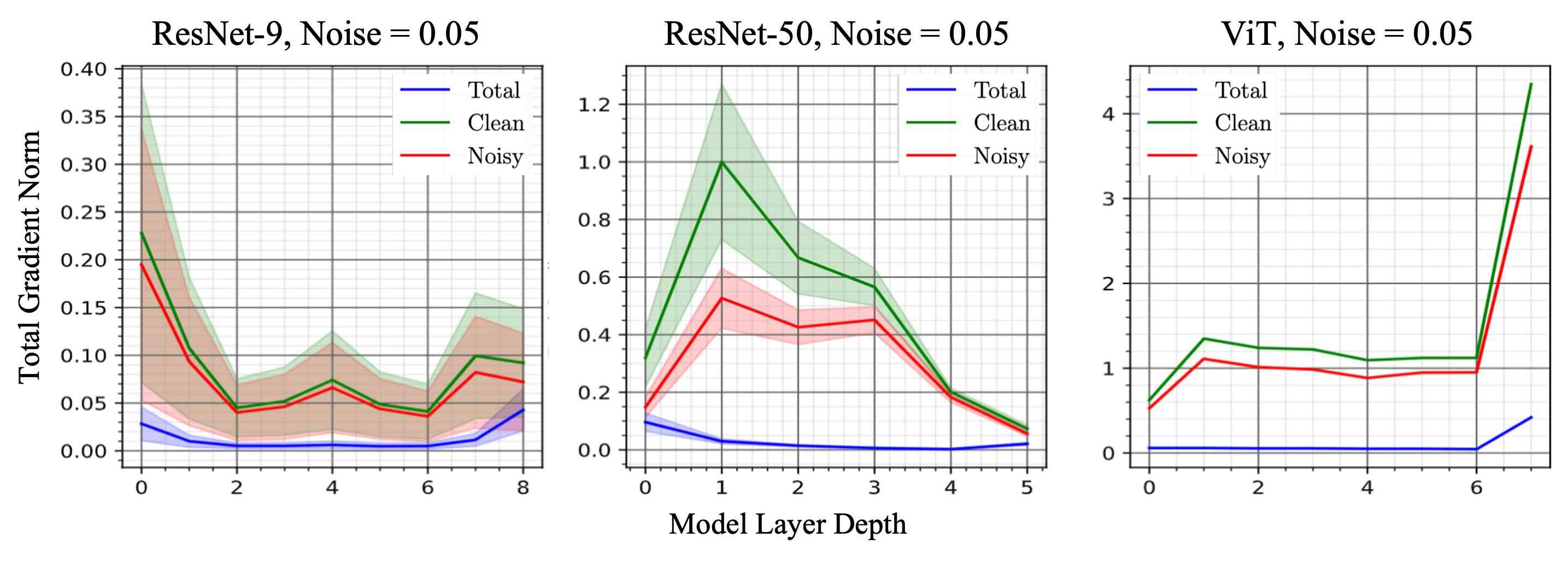}
    }\\
    \subcaptionbox{Gradient norm contribution from clean and noisy examples on when label noise is 10\% in the CIFAR-10 dataset.\label{}}{
    \includegraphics[width=0.9\textwidth]{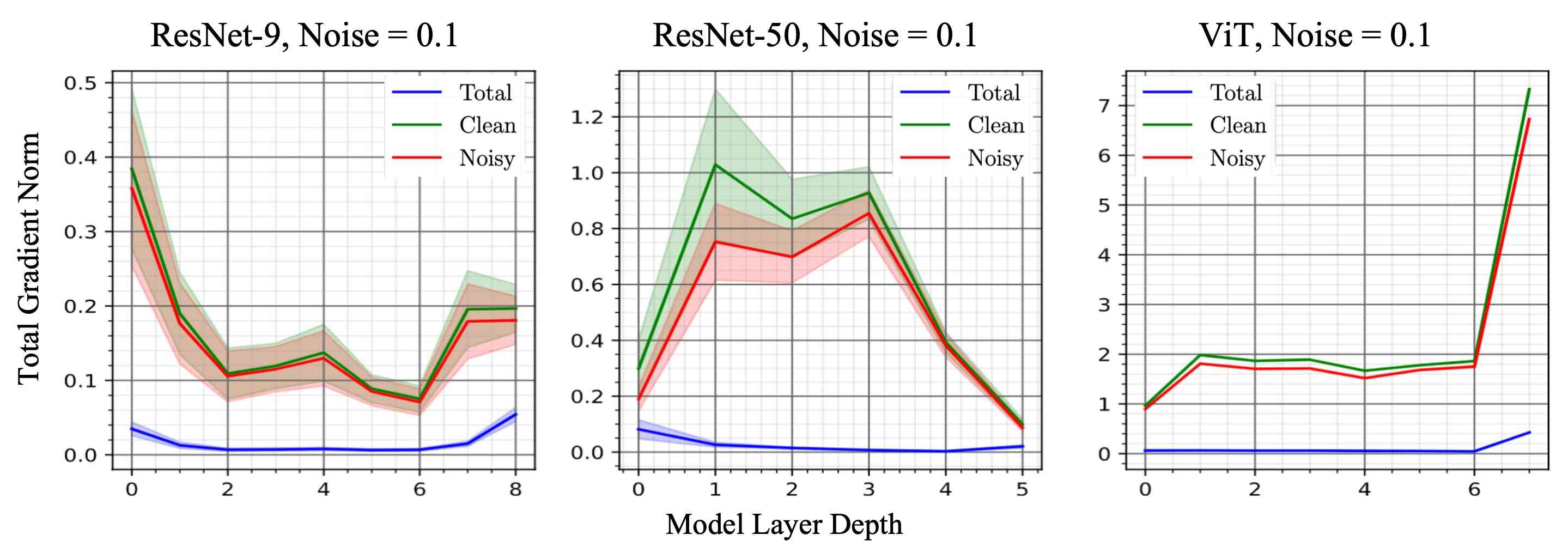}
    }\\
    \subcaptionbox{Gradient norm contribution from clean and noisy examples on when label noise is 20\% in the CIFAR-10 dataset.\label{}}{
    \includegraphics[width=0.9\textwidth]{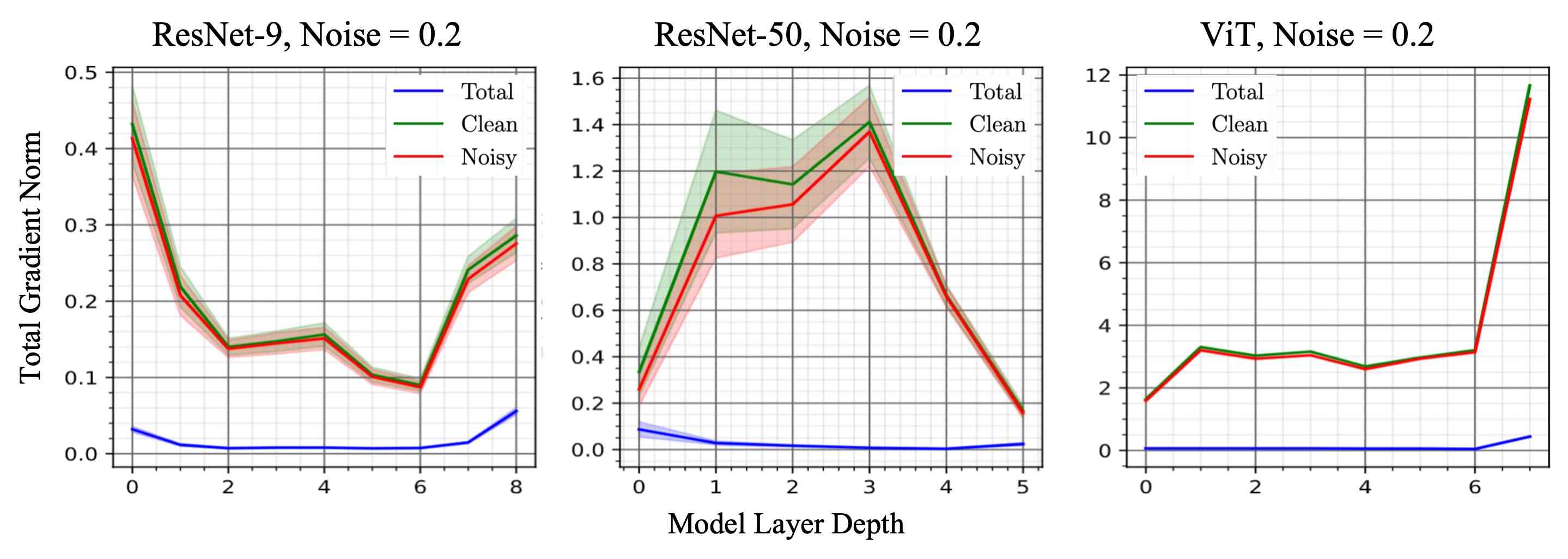}
    }
    \caption{Total gradient norm contribution (averaged over all epochs) from mislabeled and clean examples over the course of training a ResNet-9, ResNet-50, ViT model on the CIFAR-10 dataset with different proportions of label noise.}
    \label{fig:grads-contribution-cifar}
\end{figure*}

\begin{figure*}[h]
    \centering
    \subcaptionbox{Gradient norm contribution from clean and noisy examples on when label noise is 5\% in the MNIST dataset.\label{}}{
    \includegraphics[width=0.9\textwidth]{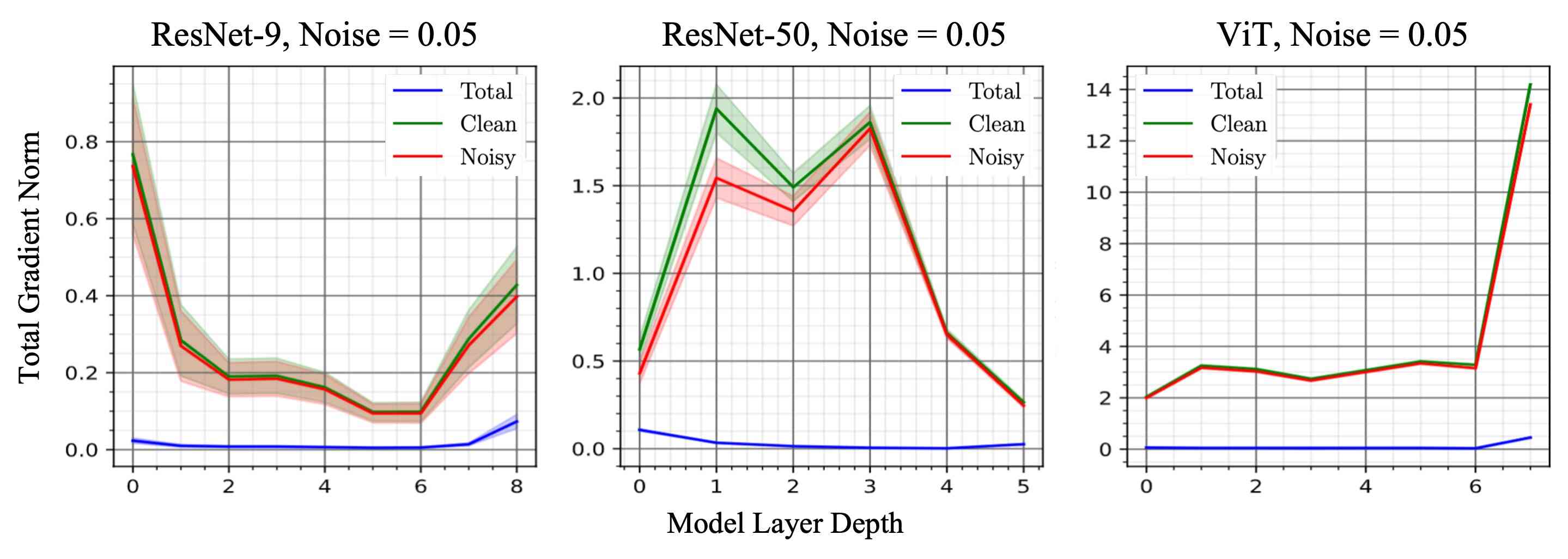}
    }\\
    \subcaptionbox{Gradient norm contribution from clean and noisy examples on when label noise is 10\% in the MNIST dataset.\label{}}{
    \includegraphics[width=0.9\textwidth]{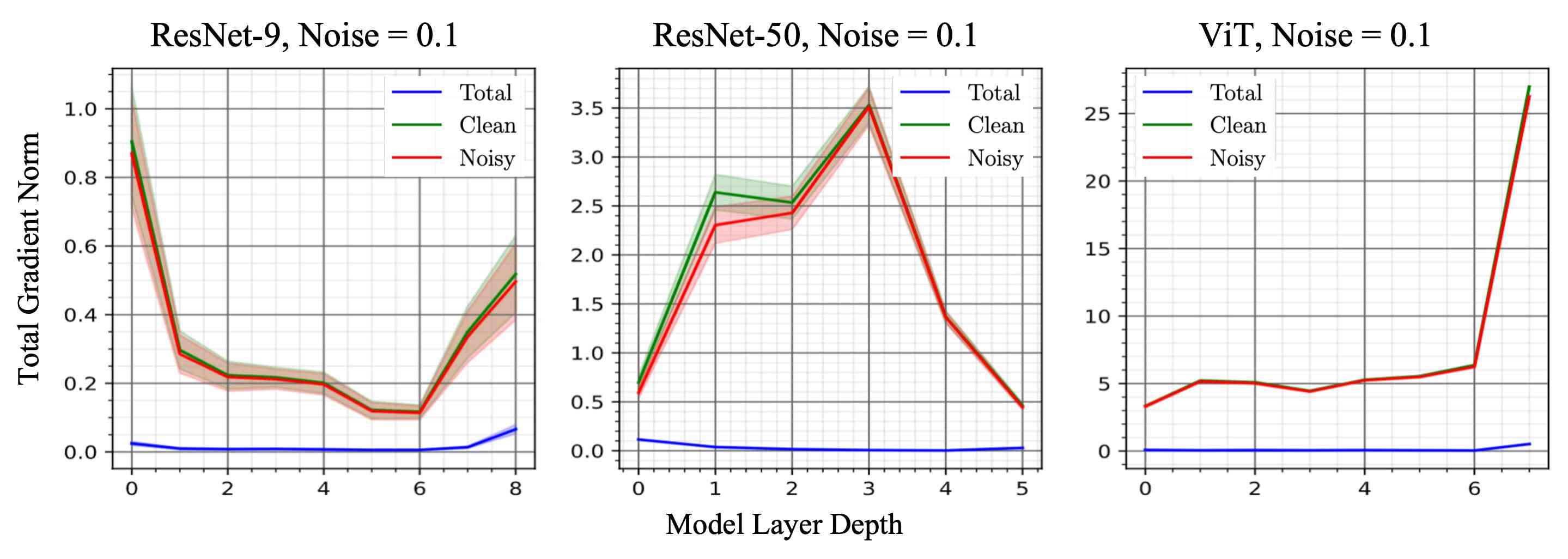}
    }\\
    \subcaptionbox{Gradient norm contribution from clean and noisy examples on when label noise is 20\% in the MNIST dataset.\label{}}{
    \includegraphics[width=0.9\textwidth]{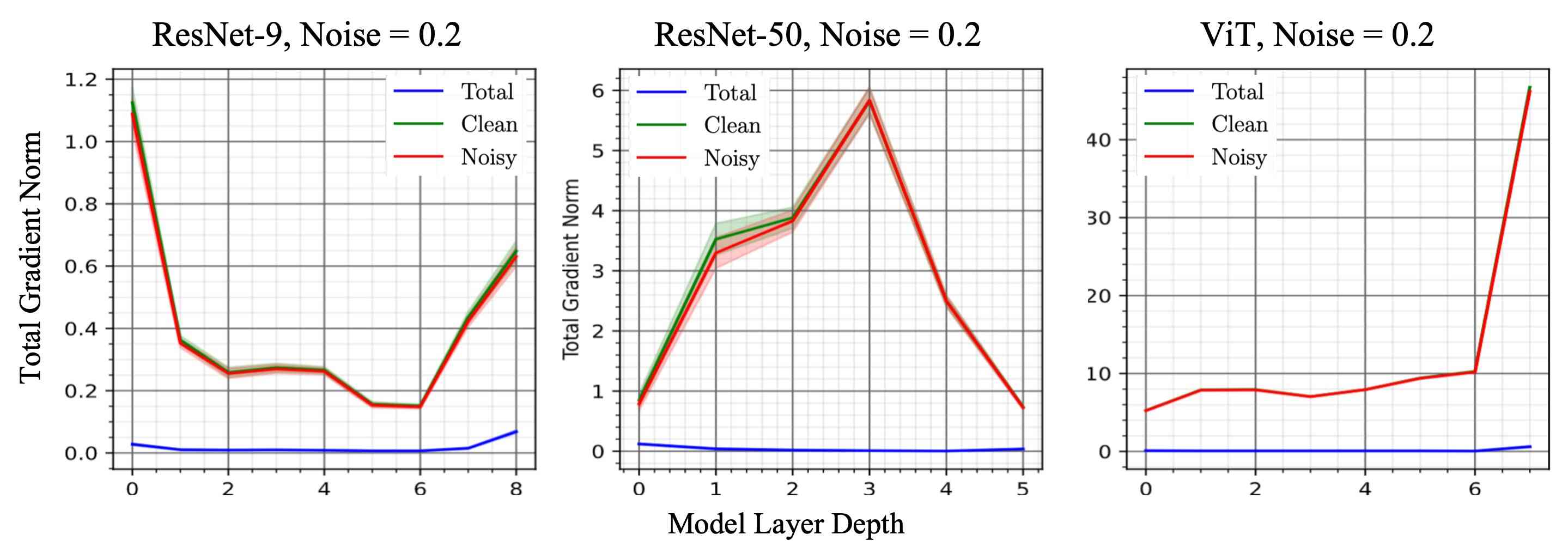}
    }
    \caption{Total gradient norm contribution (averaged over all epochs) from mislabeled and clean examples over the course of training a ResNet-9, ResNet-50, ViT model on the MNIST dataset with different proportions of label noise.}
    \label{fig:grads-contribution-mnist}
\end{figure*}

\begin{figure*}[h]
    \centering
    \subcaptionbox{Gradient norm contribution from clean and noisy examples on when label noise is 5\% in the SVHN dataset.\label{}}{
    \includegraphics[width=0.9\textwidth]{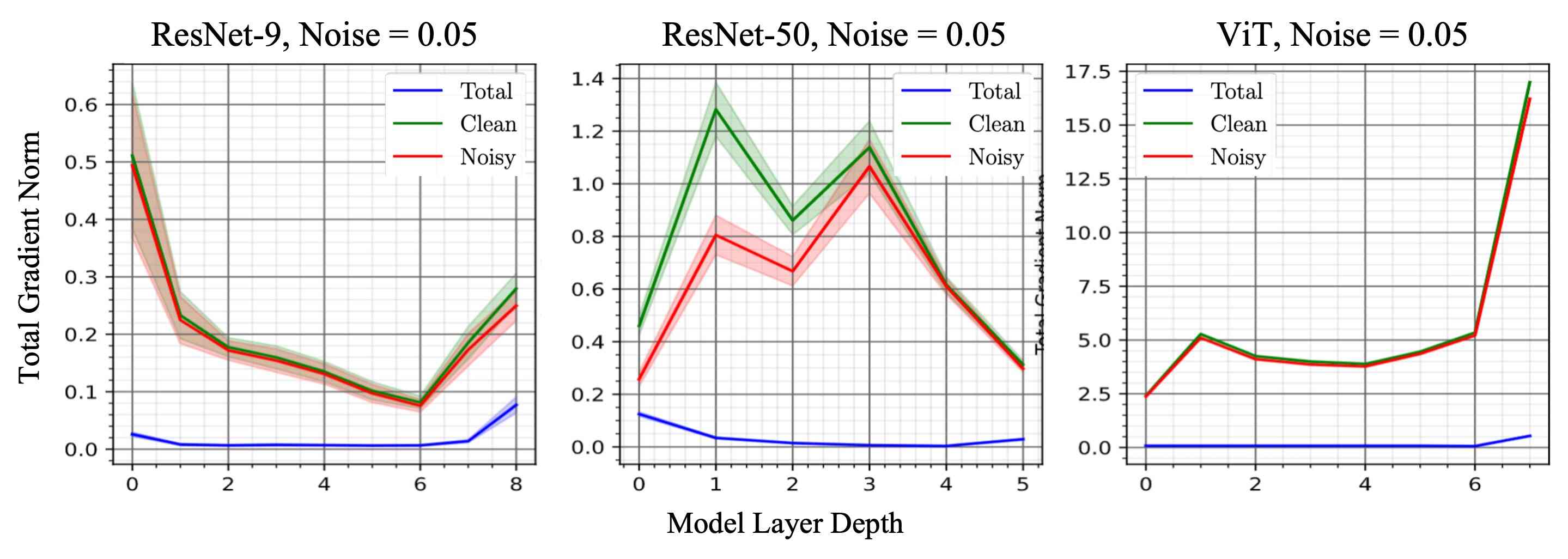}
    }\\
    \subcaptionbox{Gradient norm contribution from clean and noisy examples on when label noise is 10\% in the SVHN dataset.\label{}}{
    \includegraphics[width=0.9\textwidth]{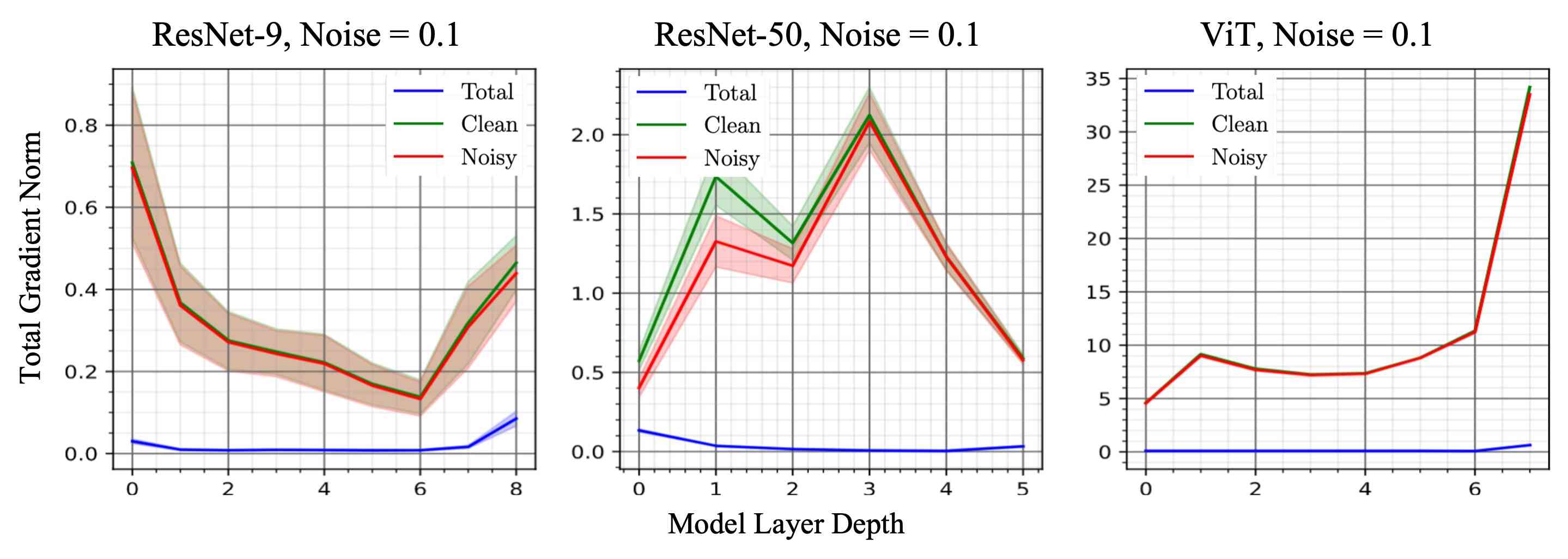}
    }\\
    \subcaptionbox{Gradient norm contribution from clean and noisy examples on when label noise is 20\% in the SVHN dataset.\label{}}{
    \includegraphics[width=0.9\textwidth]{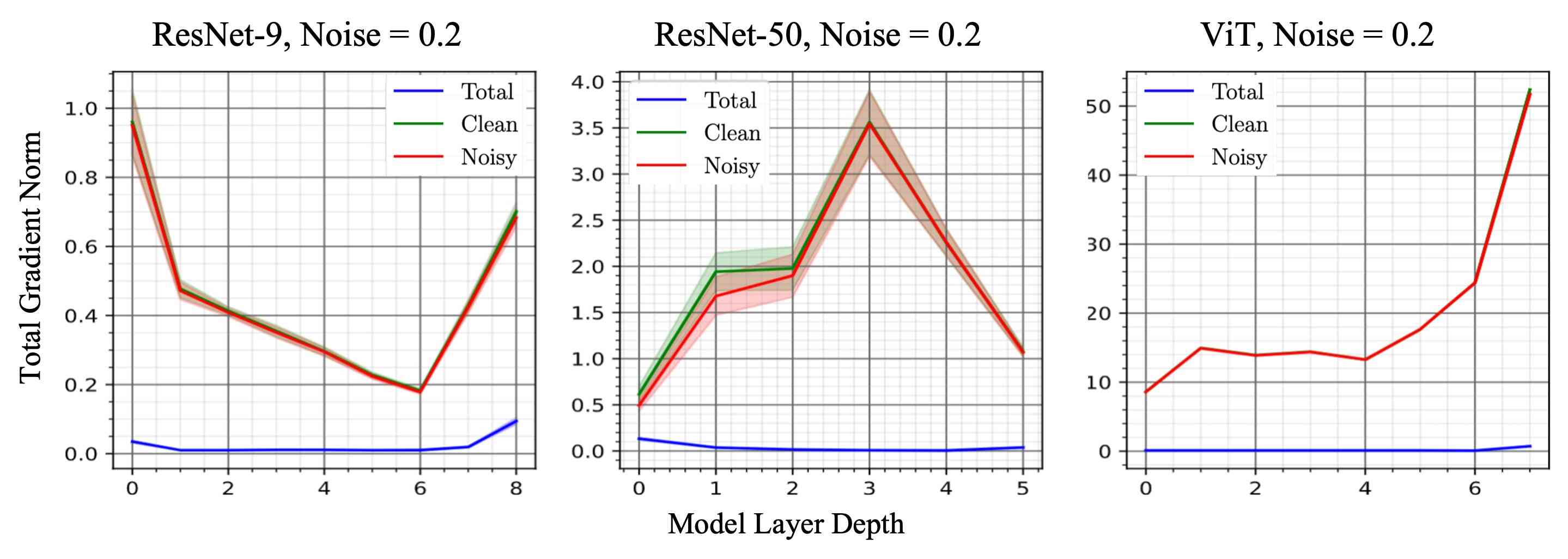}
    }
    \caption{Total gradient norm contribution (averaged over all epochs) from mislabeled and clean examples over the course of training a ResNet-9, ResNet-50, ViT model on the SVHN dataset with different proportions of label noise.}
    \label{fig:grads-contribution-svhn}
\end{figure*}

We observe that the dispersed contribution of noisy example gradients is consistent across all noise rates, models, and datasets. This is shown in Figure \ref{fig:grads-contribution-cifar}, Figure \ref{fig:grads-contribution-mnist}, and Figure \ref{fig:grads-contribution-svhn} for the CIFAR-10, MNIST, and SVHN datasets respectively.
 Moreover, we continue to consistently observe the trend that noisy example gradient contribution is nearly as high as that of clean examples even when they constitute only a small fraction of the dataset by size.

\section{Functional Assessment}
\label{app:functional}
\subsection{Layer Rewinding}

\label{app:subsubsec:rewinding}
To supplement our findings of layer rewinding on the CIFAR-10 dataset in the main paper, we provide additional results on the MNIST dataset. It can be seen from the rewinding graphs for noisy examples that the critical layers change based  on the model architecture. Moreover, when compared with the results in the main paper, the critical layers are not the same as that for CIFAR-10 dataset. (For example, in the case of the ResNet-50 model, Layer 3 was most important from a rewinding perspective, however, Layer 4 is more important in the case of the MNIST dataset.)

\begin{figure*}[t]
\centering
\subcaptionbox{CIFAR-10, Noise = 5\%.\label{}}{
  \includegraphics[width=0.68\linewidth]{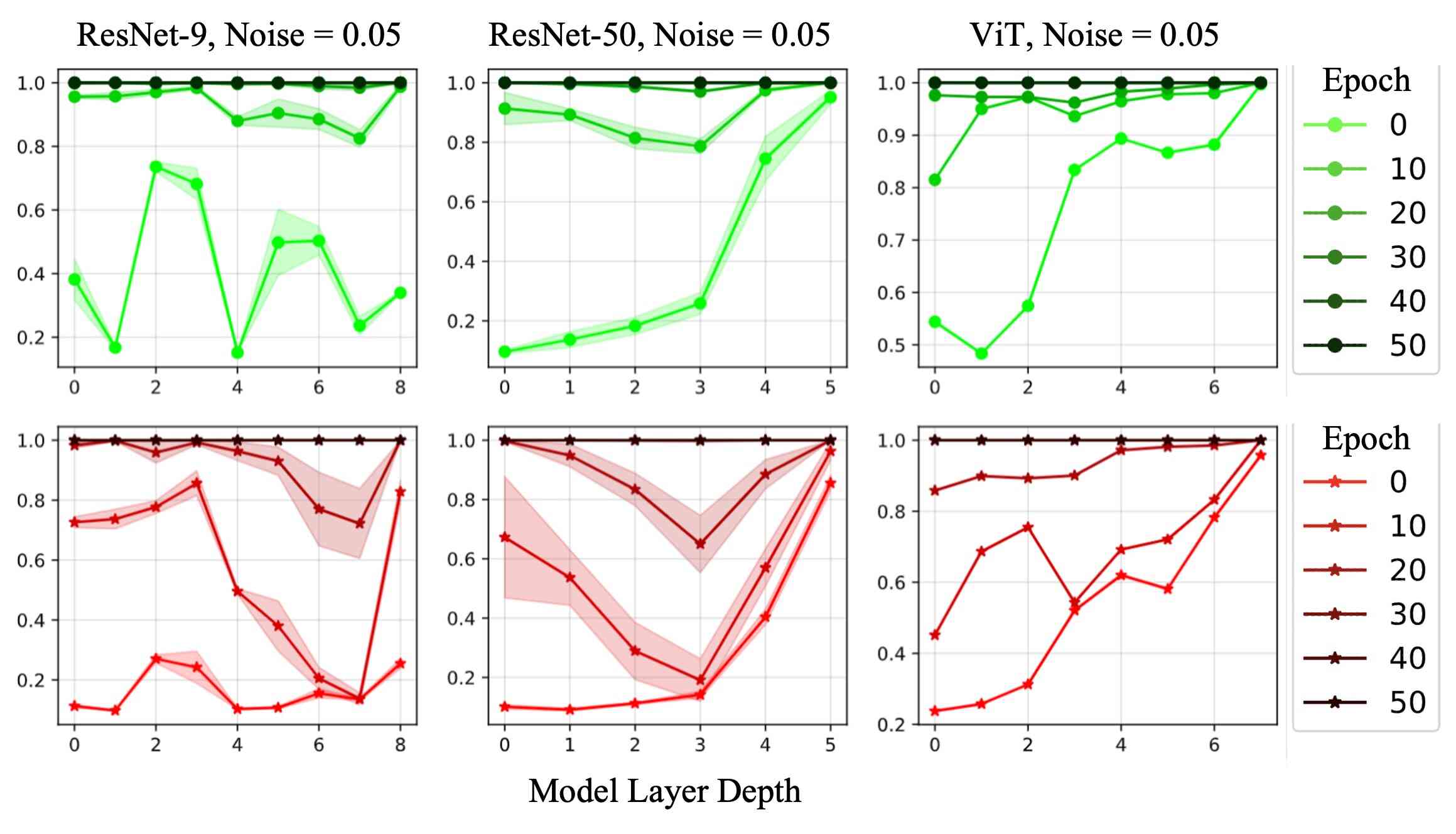}}
\\\subcaptionbox{CIFAR-10, Noise = 10\%\label{}}{
  \includegraphics[width=0.68\linewidth]{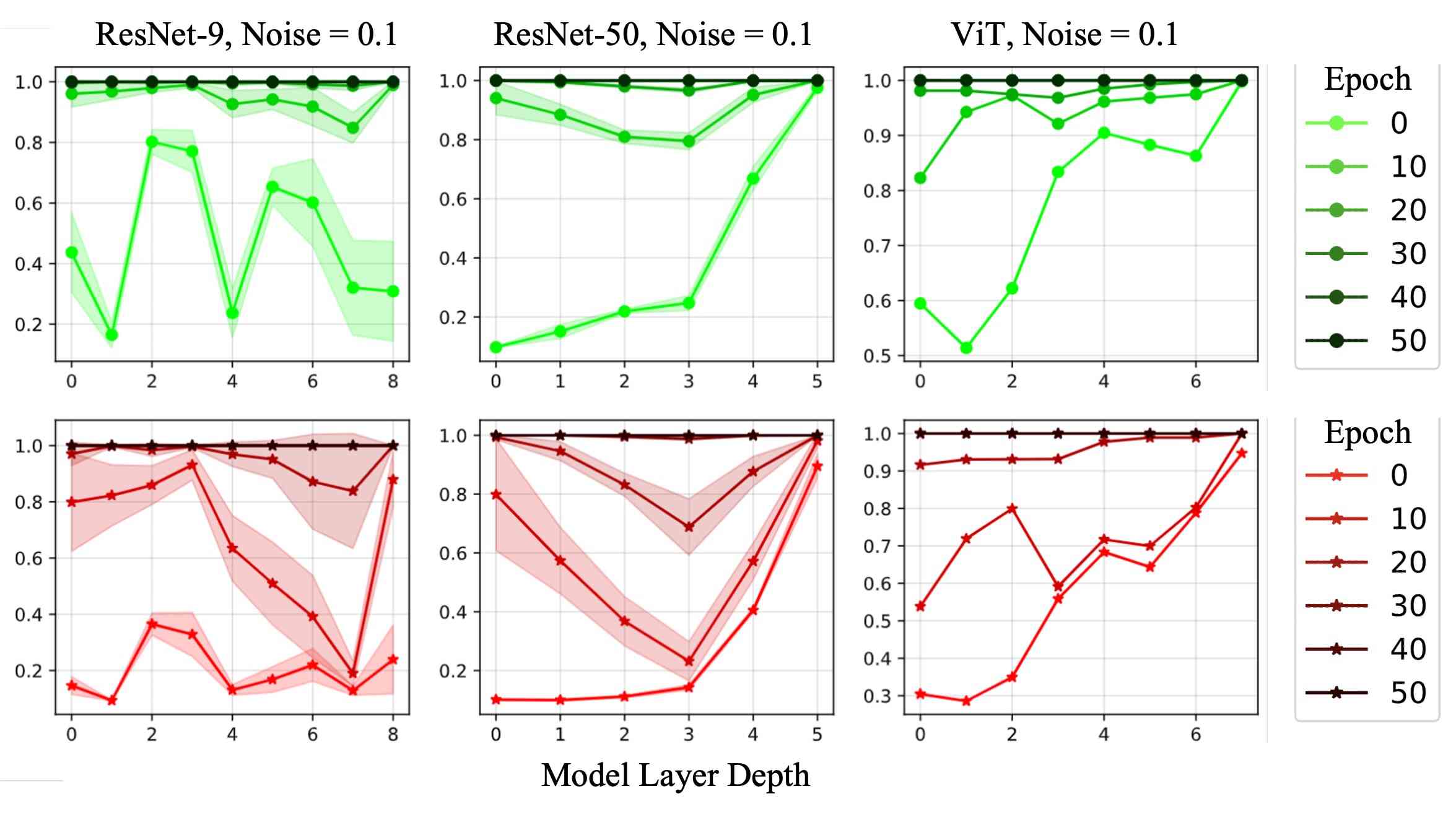}}
\\\subcaptionbox{CIFAR-10, Noise = 20\%}{
  \includegraphics[width=0.68\linewidth]{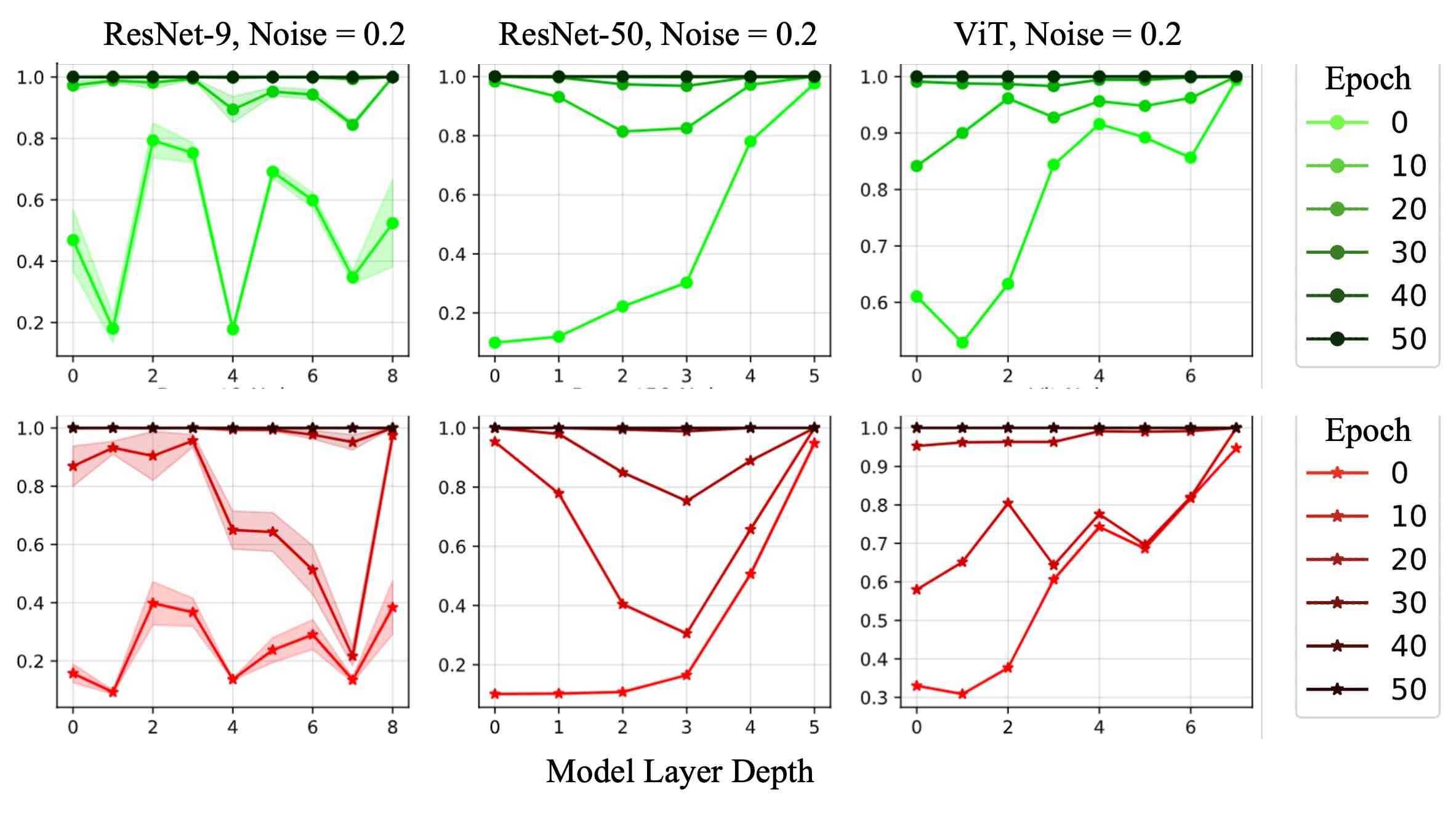}}
  \vspace{-5mm}
\caption{Change in model accuracy on rewinding individual layers to a previous training epoch for clean (top) and mislabeled
examples (bottom). Experiments are performed by training the ResNet-9, ResNet-50 and ViT models on the CIFAR-10 dataset with varying degrees of random label noise from 5\% to 20\%. Epoch 0 represents the model weights at initialization. Detailed inference is discussed in~\S~\ref{app:subsubsec:rewinding}}
\end{figure*}
\begin{figure*}[t]
\centering
\subcaptionbox{MNIST, Noise = 5\%.\label{}}{
  \includegraphics[width=0.68\linewidth]{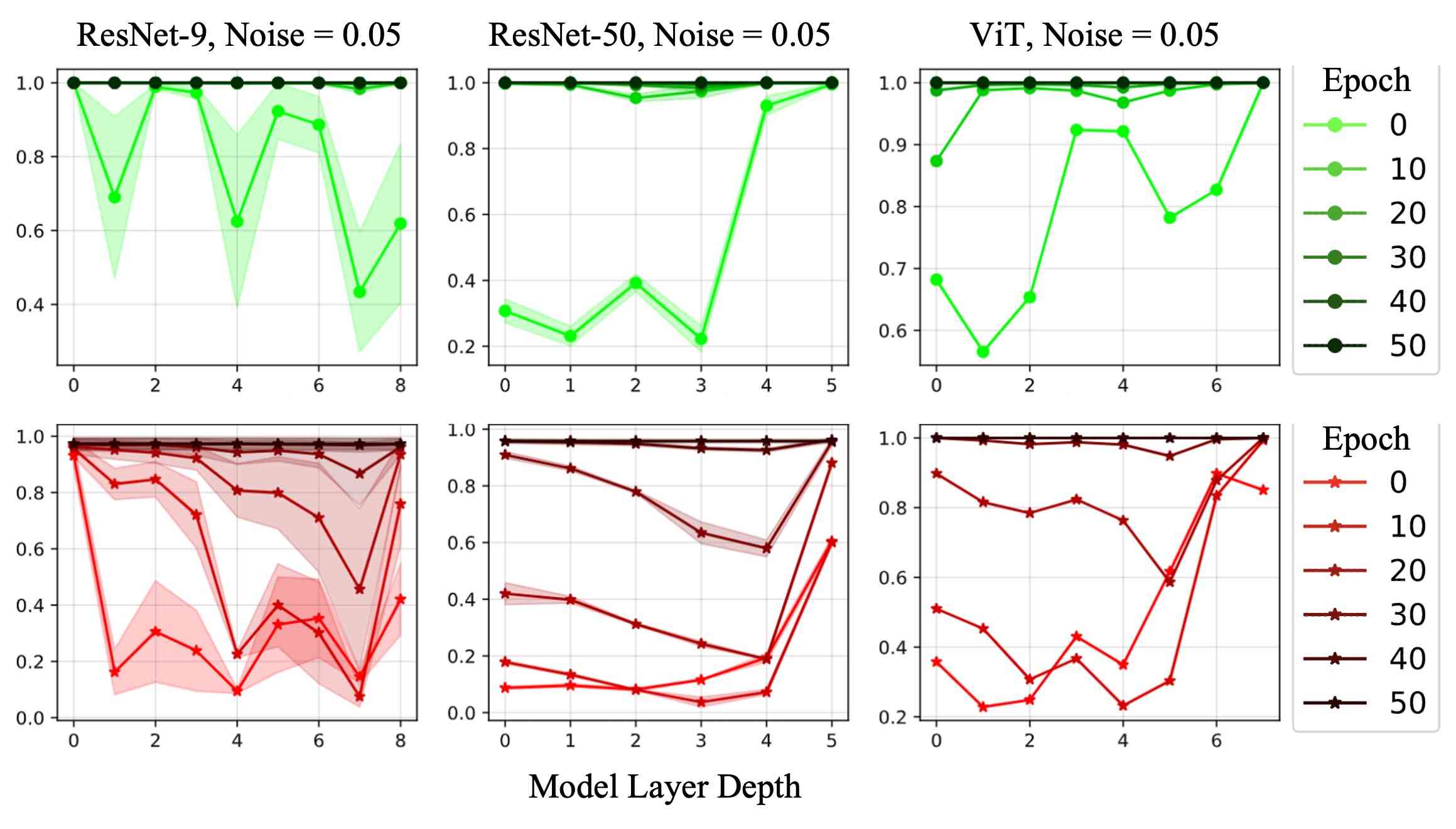}}
\\\subcaptionbox{MNIST, Noise = 10\%\label{}}{
  \includegraphics[width=0.68\linewidth]{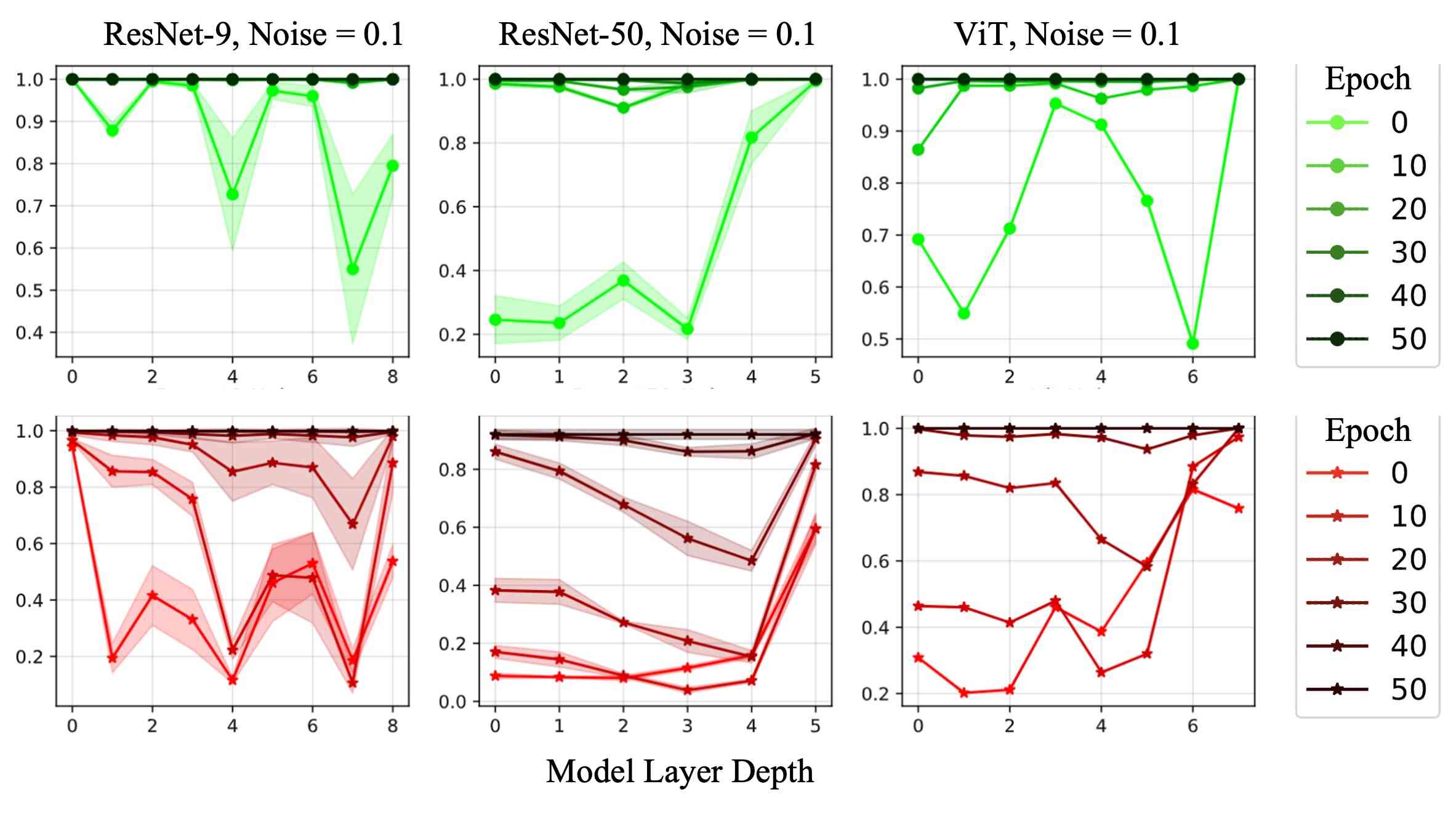}}
\\\subcaptionbox{MNIST, Noise = 20\%}{
  \includegraphics[width=0.68\linewidth]{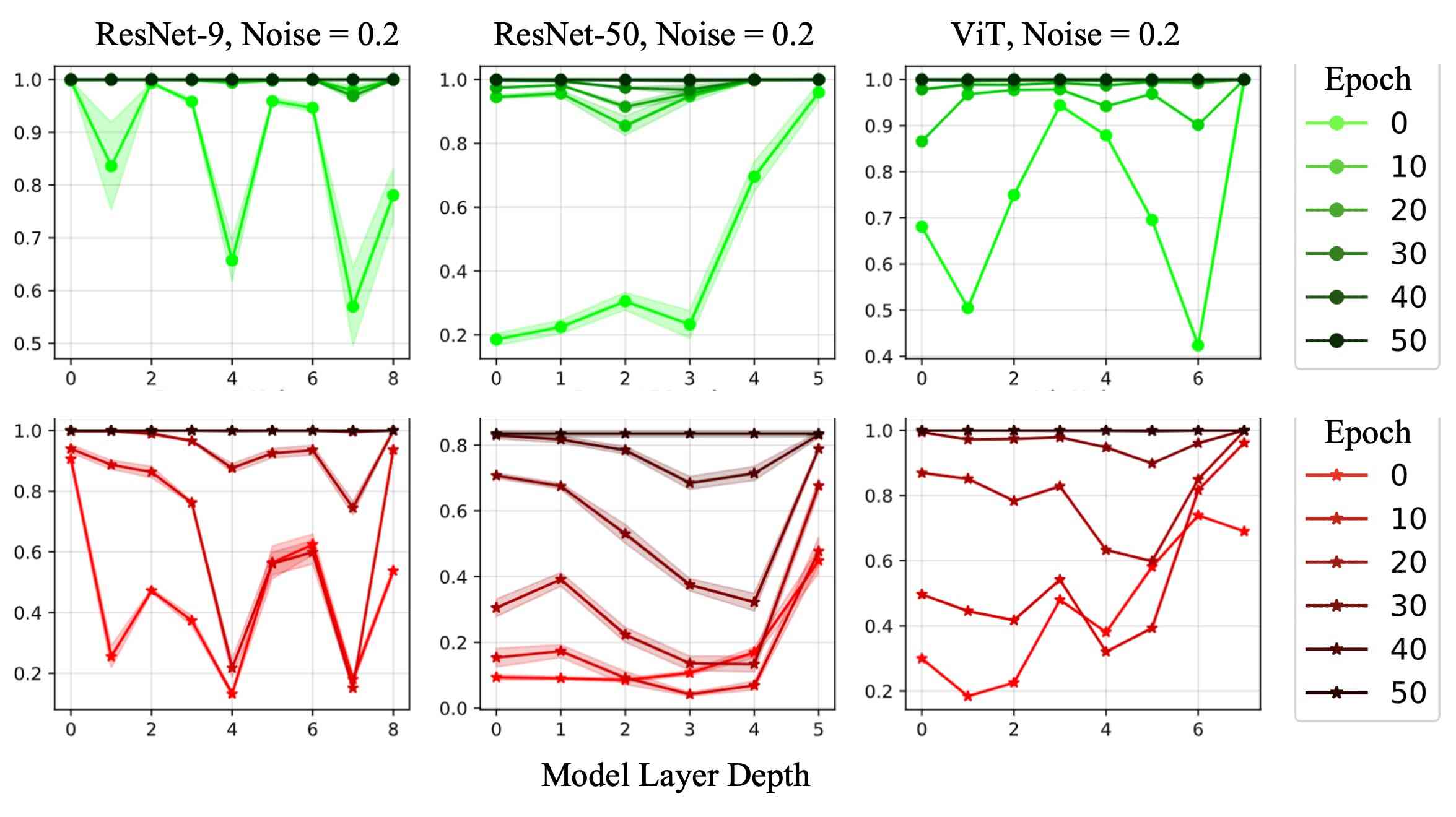}}
  \vspace{-5mm}
\caption{Change in model accuracy on rewinding individual layers to a previous training epoch for clean (top) and mislabeled
examples (bottom). Experiments are performed by training the ResNet-9, ResNet-50 and ViT models on the MNIST dataset with varying degrees of random label noise from 5\% to 20\%. Epoch 0 represents the model weights at initialization. Detailed inference is discussed in~\S~\ref{app:subsubsec:rewinding}}
\end{figure*}
\begin{figure*}[t]
\centering
\subcaptionbox{SVHN, Noise = 5\%.\label{}}{
  \includegraphics[width=0.68\linewidth]{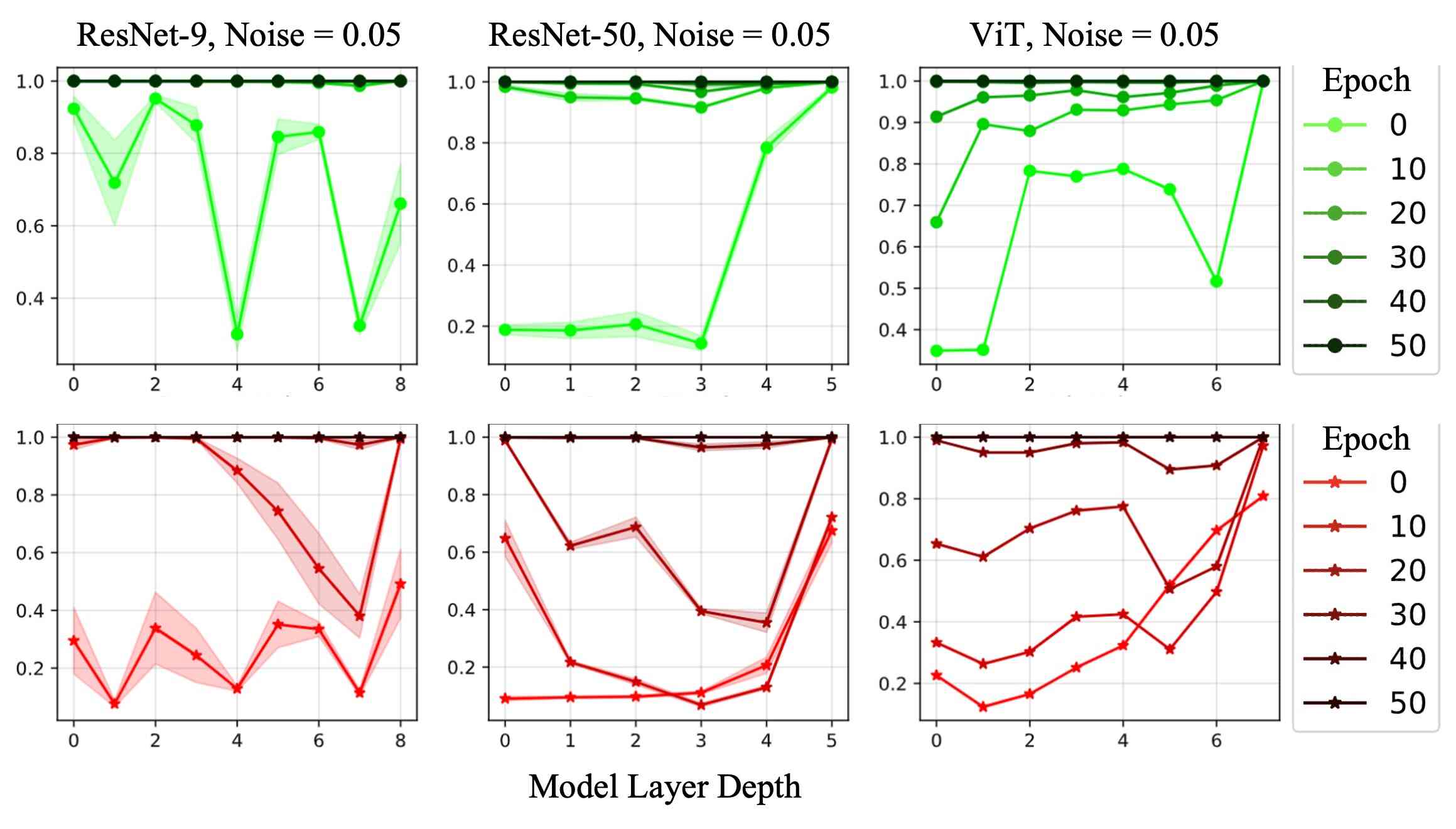}}
\\\subcaptionbox{SVHN, Noise = 10\%\label{}}{
  \includegraphics[width=0.68\linewidth]{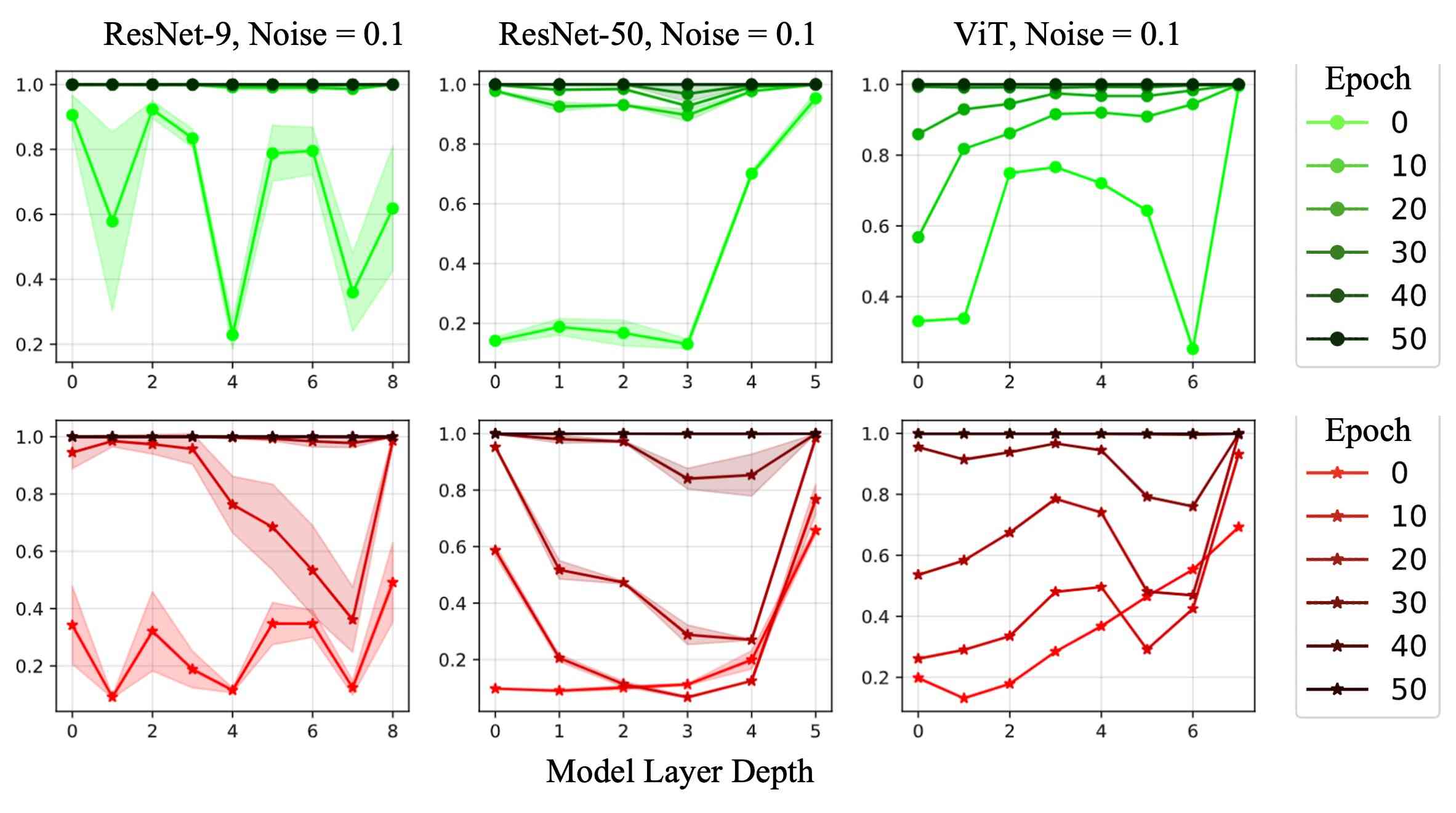}}
\\\subcaptionbox{SVHN, Noise = 20\%}{
  \includegraphics[width=0.68\linewidth]{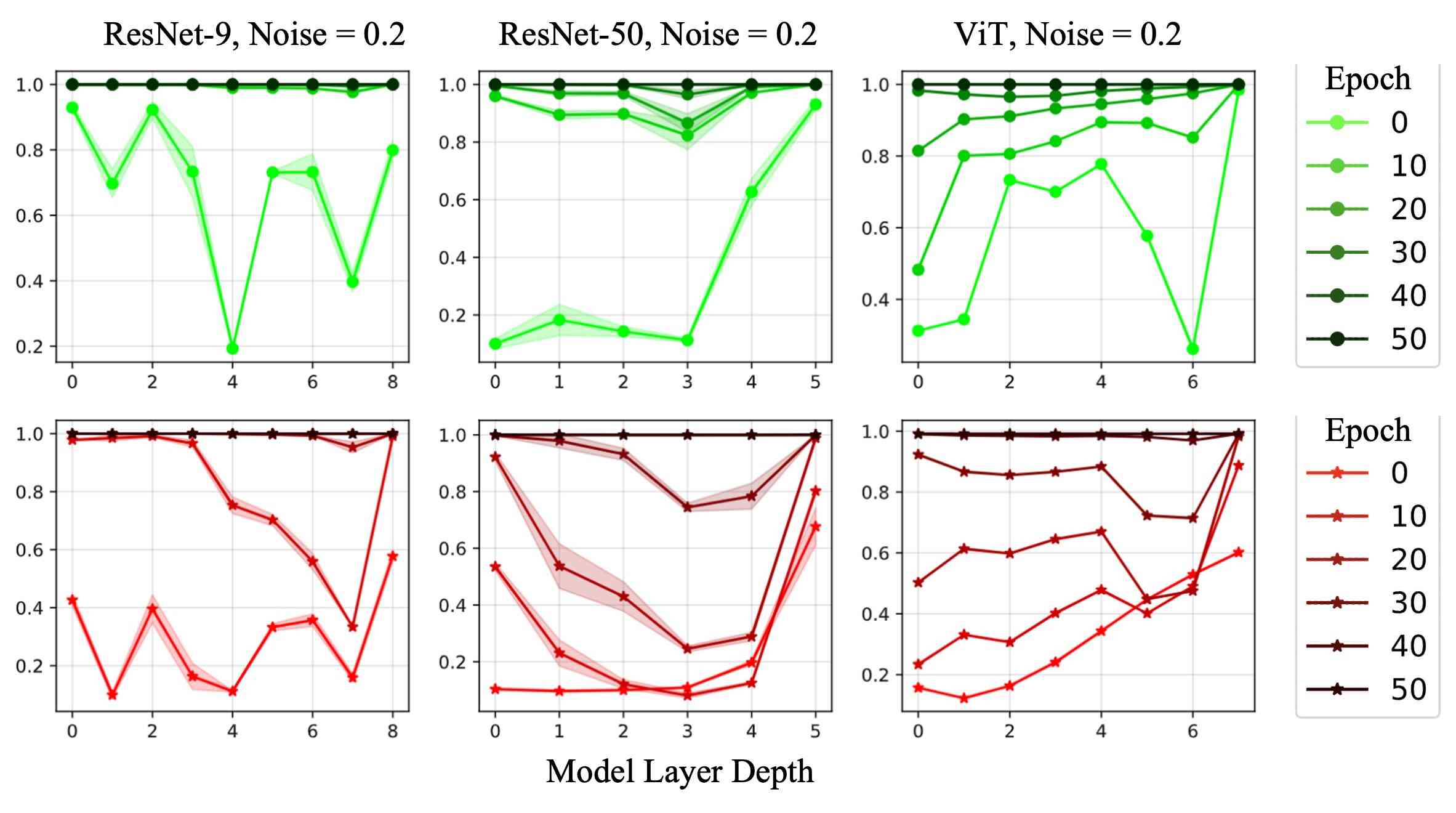}}
  \vspace{-5mm}
\caption{Change in model accuracy on rewinding individual layers to a previous training epoch for clean (top) and mislabeled
examples (bottom). Experiments are performed by training the ResNet-9, ResNet-50 and ViT models on the SVHN dataset with varying degrees of random label noise from 5\% to 20\%. Epoch 0 represents the model weights at initialization. Detailed inference is discussed in~\S~\ref{app:subsubsec:rewinding}}
 \label{app:fig:rewinding}
\end{figure*}

\subsection{Layer Retraining}
\label{app:subsubsec:retraining}

\begin{figure*}[t]
\centering
\subcaptionbox{CIFAR-10, Noise = 5\%.\label{}}{
  \includegraphics[width=0.3\linewidth]{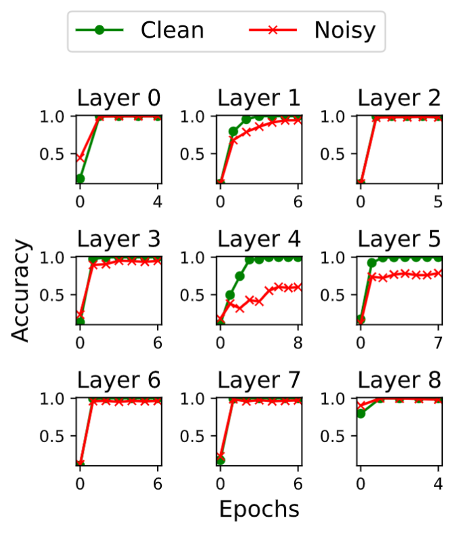}}
\subcaptionbox{CIFAR-10, Noise = 10\%\label{}}{
  \includegraphics[width=0.3\linewidth]{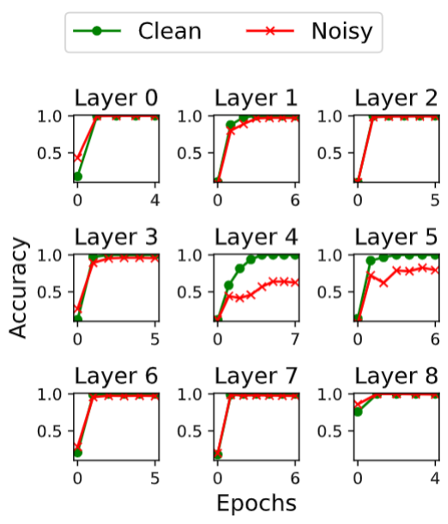}}
\subcaptionbox{CIFAR-10, Noise = 20\%}{
  \includegraphics[width=0.3\linewidth]{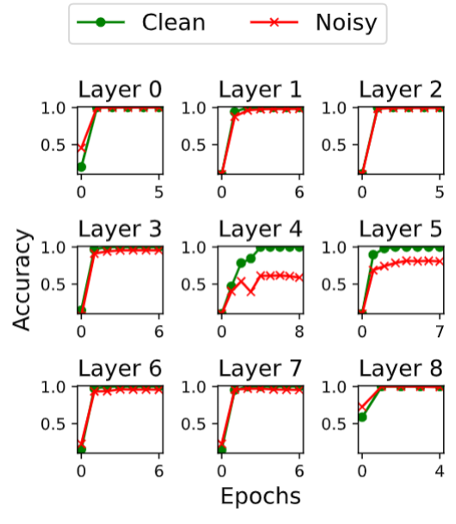}}
\\
\subcaptionbox{MNIST, Noise = 5\%.\label{}}{
  \includegraphics[width=0.3\linewidth]{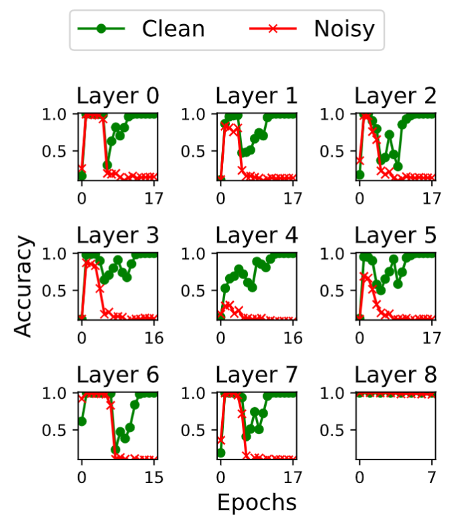}}
\subcaptionbox{MNIST, Noise = 10\%\label{}}{
  \includegraphics[width=0.3\linewidth]{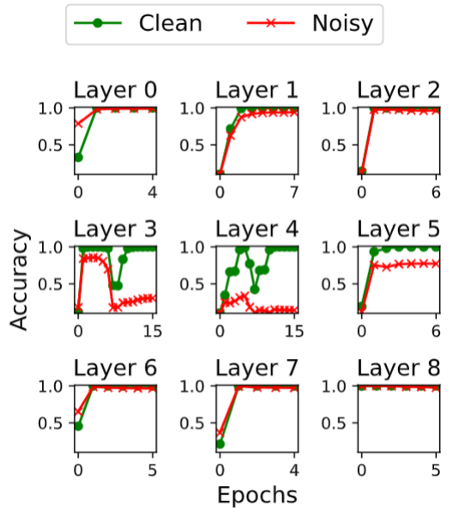}}
\subcaptionbox{MNIST, Noise = 20\%}{
  \includegraphics[width=0.3\linewidth]{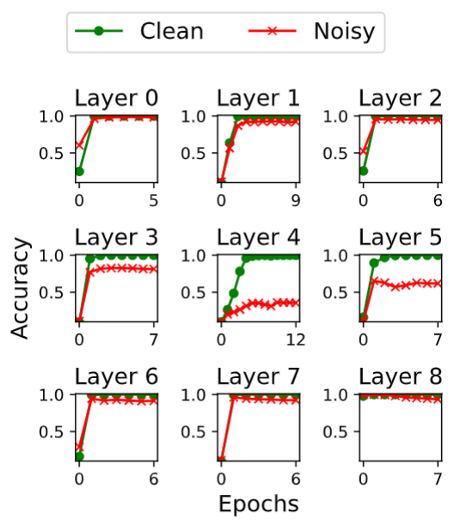}}
\\
\subcaptionbox{SVHN, Noise = 5\%.\label{}}{
  \includegraphics[width=0.3\linewidth]{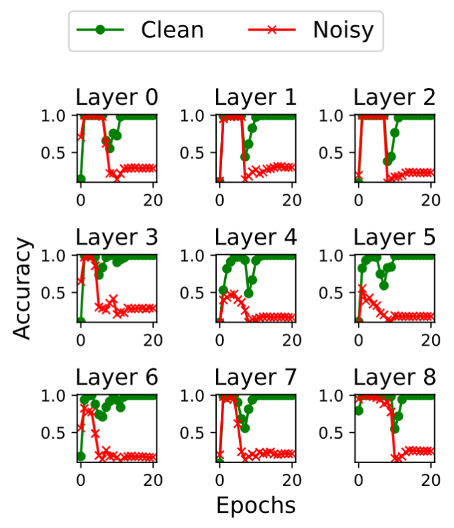}}
\subcaptionbox{SVHN, Noise = 10\%\label{}}{
  \includegraphics[width=0.3\linewidth]{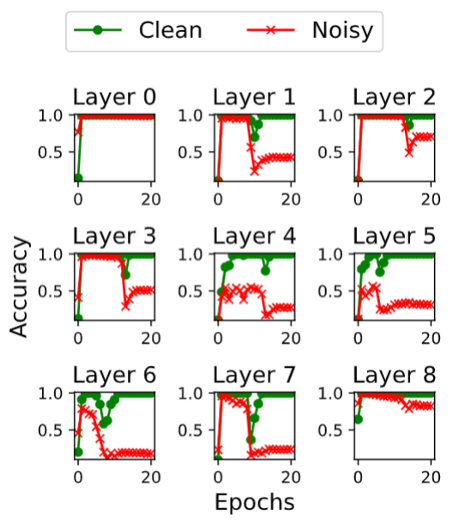}}
\subcaptionbox{SVHN, Noise = 20\%}{
  \includegraphics[width=0.3\linewidth]{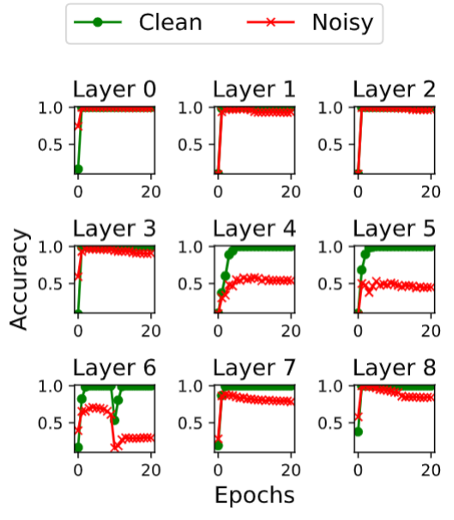}}
\caption{We trained ResNet-9 models at varying percentages of random noise. The graphs show the change in accuracy of clean and noisy examples when retraining different layers (from initialization) only on clean examples. Detailed inference is discussed in~\S~\ref{app:subsubsec:retraining}}
 \label{app:fig:retraining}
\end{figure*}

\paragraph{Retraining for Memorization of Mislabeled Examples}
We present the layers for layer retraining on the CIFAR-10, MNIST, and SVHN datasets at three different noise levels of 5\%, 10\%, and 20\% label noise in Figure~\ref{app:fig:retraining}. It may be observed that even though the model is able to achieve high accuracy on noisy examples at an intermediate epoch during the retraining schedule, the model eventually converges to an alternate minima where the performance on mislabeled examples is close to random. This once again suggests that the existence of a high accuracy point during the training schedule is sufficient to show that the particular layer was not important for memorization, but the converse is not true.

\begin{figure*}
\centering
  \includegraphics[width=0.8\linewidth]{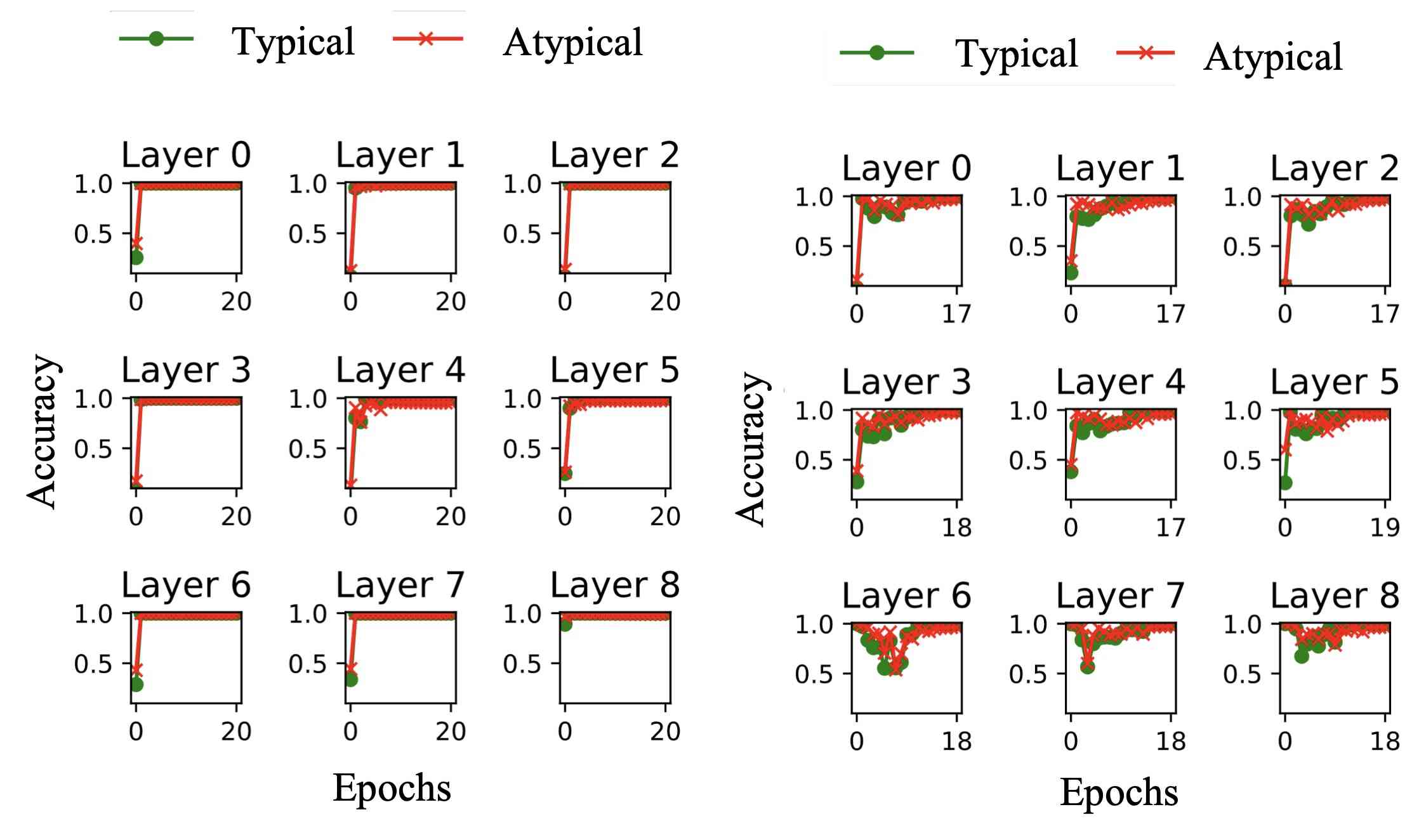}
  \caption{We trained ResNet-9 models on the MNIST and CIFAR-10 datasets with no label noise. We then retrained each layer (from initialization) only on typical examples (C-score $>$ 0.5) and atypical examples (C-score $<$ 0.5). The graphs show the change in accuracy of typical and atypical examples. Detailed inference is discussed in~\S~\ref{app:subsubsec:retraining}}
  \label{app:fig:retraining-atypical}
\end{figure*}

\paragraph{Retraining for Memorization of Atypical Examples}
Using the definition of consistency score (C-score), we categorize all examples in the MNIST and CIFAR-10 datasets as atypical if the C-score is less than 0.5. We retrain each layer of a ResNet-9 model trained on the MNIST and CIFAR-10 datasets with no label noise and analyze the learning dynamics of typical and atypical examples. The results are shown in Figure~\ref{app:fig:retraining-atypical}. We observe that the model is able to achieve high accuracy on atypical examples at an intermediate epoch during the retraining schedule. This once again suggests that training on typical examples is sufficient to achieve high accuracy on atypical examples. Moreover, unlike the case of  mislabeled examples where layers 4 and 5 were important for memorization, we observe that none of the layers are important for memorization of atypical examples. This suggests that the memorization of atypical examples is a more dispersed phenomenon than the memorization of mislabeled examples.

\end{document}